%% file: colm2025_conference.tex
\documentclass{article} %
\usepackage[final]{colm2025_conference}
\usepackage{stroop}
\usepackage{twemojis}

\usepackage{microtype}
\usepackage{hyperref}
\usepackage{url}
\usepackage{booktabs}

\usepackage{lineno}

\definecolor{darkblue}{rgb}{0, 0, 0.5}
\hypersetup{colorlinks=true, citecolor=darkblue, linkcolor=darkblue, urlcolor=darkblue}

\title{Control the Temperature: Selective Sampling for Diverse and High-Quality LLM Outputs}

\newcommand{\affone}{1}
\newcommand{\afftwo}{2}

\author{
\noindent\begin{tabular}{@{}l@{}}
Sergey Troshin\thanks{Authors contributed equally; correspondence to \texttt{serj.troshin@proton.me}}\,\,\,$^{,\affone}$,   %
Wafaa Mohammed$^{\ast,\affone}$, 
Yan Meng$^{\ast, \affone}$, \\
{Christof Monz}$^{\affone}$, 
{Antske Fokkens}$^{\afftwo}$,
{Vlad Niculae}$^{\affone}$
\end{tabular} \\
$^{\affone}$\,Language Technology Lab, University of Amsterdam \\ 
$^{\afftwo}$\,Computational Linguistics and Text Mining Lab, Vrije Universiteit Amsterdam
}

\newcommand{\risk}{sampling risk\,}

\begin{document}

\ifcolmsubmission
\linenumbers
\fi

\maketitle

\begin{abstract}
Diversity is an essential metric for evaluating the creativity of outputs generated by language models.
Temperature-based sampling is a common strategy to increase diversity. 
However, for tasks that require high precision, e.g., mathematical reasoning, uncontrolled high temperature sampling, e.g., min-$p$ or top-$p$, degrades reasoning quality. 
We demonstrate that the loss of accuracy is caused by sampling incorrect continuations in sensitive decoding positions. 
To address this, in this paper, we propose \textbf{selective sampling}, a method that dynamically switches between greedy and high-temperature sampling based on a sampling risk metric. 
This risk metric estimates the likelihood of output errors when applying high-temperature sampling on the current token position. 
To predict sampling risk, we train a lightweight classifier on a small subset of verifiable problems. 
The trained classifier can be integrated with the base language model with minimal latency overhead. 
Experiments on mathematical reasoning tasks demonstrate that selective sampling enhances the quality-diversity trade-off, even in high-temperature settings.

\end{abstract}

\input{sections/introduction}

\input{sections/preliminaries}

\input{sections/sampling_risk_analysis}

\input{sections/methodology}

\input{sections/experiments}

\input{sections/results}

\input{sections/related_work}

\input{sections/conclusion}

\section{Limitations}
Our work exhibits a few limitations.
First, while our classifier for selective sampling is easy to implement, the classifier is model-dependent, which makes our approach not directly transferable between different models compared to the baseline methods. 
Second, we focus on the mathematical reasoning tasks, because they provide a simple way to estimate correctness using verifiable rewards (accuracy of the final answers). 
We do not incorporate the correctness of the CoTs as part of the quality metric, which would be much harder to evaluate. 
We are looking forward to exploring more open-ended and creative writing tasks in the future, with more subtle variations in sampling risks. 
Third, for the diversity evaluation, we only use a commonly used and intuitive n-gram-based diversity measure.

\section*{Reproducibility Statement}
Our code is based on open-source software libraries for training and evaluation of LLMs: vLLM \citep{kwon2023efficient}, Evaluation Harness \citep{eval-harness}, Hugging Face \citep{wolf2020huggingfacestransformersstateoftheartnatural}. We conduct the experiments on commonly used datasets available via Hugging Face \citep{wolf2020huggingfacestransformersstateoftheartnatural}. We use the prompt formats defined in Evaluation Harness \citep{eval-harness}. Regarding the used datasets, the GSM8k dataset \citep{cobbe2021gsm8k} is openly available under the MIT license. Minerva MATH Prealgebra subset is part of the MATH dataset \citep{hendrycksmath2021}, and by using it, we refer to the Fair Use case discussed in \citet{hendrycksmath2021}. Due to a recent controversy about the copyright status of the MATH dataset, we can only release data or models trained on GSM8K and GSM-symbolic. We will revisit the release of the Minerva MATH part of the data if this situation changes. This may affect the reproducibility of part of our experiments (mainly those in which we compare our classifier train on top of the hidden states with its n-gram ablation). We call for more alternative evaluation datasets designed with the intention of being used for research and released with open licenses.

\section*{Ethics Statement}
 While we mitigate the degradation in quality for high temperature sampling, we have not measured whether this mitigation also affects toxicity or other undesirable properties of text. We therefore recommend the same care that users should take with any other LLM generation systems. Our method depends on the definition of task reward, which does not account for all important characteristics users might value. Since sampling risk is estimated with a classifier, it could possibly be that by suppressing certain outputs our method affects the fairness and broader diversity of the outputs.

\section*{Acknowledgments}
This publication is part of the project VI.Veni.212.228 of the research program
`Veni', which is financed by the Dutch Research Council (NWO); and is part of
‘Hybrid Intelligence: augmenting human intellect’
(https://hybrid-intelligence-centre.nl) with project number 024.004.022 of the
research program `Gravitation' which is (partly) financed by the Dutch Research
Council (NWO). 
Moreover, it is funded in part by the Netherlands Organization for Scientific Research (NWO) under the project number VI.C.192.080.
It is also a part of the UTTER project,
supported by the European Union’s Horizon Europe research and innovation programme via grant
agreement 101070631.

We thank Wilker Aziz, members of LTL and CLTL labs, and the reviewers for fruitful discussions and feedback.

\bibliography{anthology,colm2025_conference}
\bibliographystyle{colm2025_conference}

\appendix
\section*{Appendix}

\input{sections/app}

\end{document}

%% file: sections/introduction.tex
\section{Introduction}

\input{sections/introduction_yan}

%% file: sections/introduction_yan.tex
Recently, large language models (LLMs) have demonstrated unprecedented capabilities in mathematical reasoning tasks by efficiently harnessing task-specific rewards \citep{openai2024gpt4technicalreport, Dubey2024TheL3, Yang2024Qwen25TR,deepseekai2025deepseekv3technicalreport}. 
To maintain high reasoning accuracy, prior work has primarily focused on deterministic decoding, e.g., greedy decoding, which generates outputs with the highest probability. 
However, this approach often leads to pathological solutions by over-optimizing for a single objective, i.e., accuracy, at the expense of other desirable properties \citep{hashimoto-etal-2019-unifying}. 
In particular, deterministic decoding tends to reduce the diversity in generated outputs \citep{le-bronnec-etal-2024-exploring, kirk2024understandingeffectsrlhfllm}. 
This is problematic because human preferences are not always aligned with those of greedy generations due to a lack of diversity and fluency \citep{zhang-etal-2021-trading, Holtzman2020TheCuriousCase}.

Diversity is important for controllability in large language models: by having access to a probabilistic sampler, a user can first generate multiple outputs, and then select the best ones using a task-specific metric \citep{bestofn_summarization}; a user can improve parallel exploration of potential solutions \citep{shunyu_tree_of_thoughts}, or control the generation process from a black-box sampler according to additional constraints \citep{mudgal2024controlled, deng-raffel-2023-reward, troshin2024efficientcontrolledlanguagegeneration}.
Temperature sampling is a common strategy to increase creativity and diversity in many LLM inference frameworks such as vLLM \citep{kwon2023efficient} or Hugging Face \citep{wolf2020huggingfacestransformersstateoftheartnatural}. 
However, it often comes at the cost of lower task accuracy compared to deterministic decoding \citep{shi-etal-2024-thorough}.  
This trade-off has motivated research on how to narrow the quality gap between deterministic and temperature sampling \citep{Holtzman2020TheCuriousCase, basu2021mirostat, hewitt-etal-2022-truncation, nguyen2024turningheatminpsampling}.

Previous sampling methods, such as top-$p$ and min-$p$ sampling, truncate the next-token distribution based on model confidence in order to improve the quality-diversity trade-off in language generation. 
These sampling methods prioritize high-likelihood candidates to improve output precision \citep{meister-etal-2023-efficacy}. 
However, one limitation of the truncation sampling approaches is that they solely rely on model confidence to sample potential candidate tokens. 
While the shape of the model output distribution over candidate tokens can represent uncertainty, it is hard to distinguish between different types of uncertainty \citep{baan2023uncertaintynaturallanguagegeneration}, 
namely between variability due to numerous plausible continuations
\citep {giulianelli-etal-2023-comes} 
and uncertainty about what the correct answer is.
For example, if a model assigns high probability to two different answers, are both of these answers plausible, or is the model not confident which one is correct in a given context?
As we demonstrate in \cref{sec:analysis}, sampling at certain decoding time steps can lead to incorrect continuations in cases where the greedy continuations are correct. 
This shows that it is important to adopt different decoding strategies across decoding positions to balance quality and diversity.

To balance both the diversity and quality of LLMs output, we propose \textbf{selective sampling}, a method that dynamically switches between greedy and temperature sampling based on a \textit{sampling risk} metric. 
This metric measures the likelihood of output errors for a given decoding timestep when applying temperature sampling to it. 
We train a lightweight classifier on a small subset of verifiable problems to predict sampling risk. Our approach preserves the original model outputs and is easy to implement, with the classifier integrating into the base language model with minimal latency overhead. 
We empirically demonstrate that temperature sampling with our classifier results in a better quality-diversity trade-off compared to commonly used truncation-based and entropy-based sampling methods, both under standard and high-temperature settings. 

The structure of our paper is as follows: In \Cref{sec:preliminaries} we review the fundamental theoretical concepts behind the sampling methods explored in our study. 
In \Cref{sec:analysis}, we highlight the gap we aim to fill by analyzing when existing sampling methods fail. 
In \Cref{sec:method}, we detail the intuition and implementation of the selective sampling method. 
In \Cref{sec:exps} we outline the details of our experimental setup, including models, tasks, and evaluation metrics. 
Then, in \Cref{sec:results}, we present and discuss our results and findings. 
We describe related work in \Cref{sec:relwork}. Finally, we summarize our conclusions in \Cref{sec:conclusion}.

%% file: sections/preliminaries.tex
\section{Preliminaries}
\label{sec:preliminaries}
In this section, we cover key concepts for sampling mechanisms, including temperature sampling and its variants.

\subsection{Temperature Sampling}
At each step of decoding, the base model observes an already generated prefix $x$, and predicts the logits 
 $z_{\text{LM}}(\cdot | x) \in \mathbb{R}^{|V|}$. Logits are then used to form a sampling distribution $p(v|x)$, which may involve filtering, temperature, and other modifications \citep{nguyen2024turningheatminpsampling, Holtzman2020TheCuriousCase, basu2021mirostat, hewitt-etal-2022-truncation}. 
 In particular, \emph{temperature sampling} uses a scalar parameter to rescale the probabilities to make the distribution more peaked (low temperature) or flatter (high temperature). 
 The next token is then sampled from a categorical distribution defined by $p(v|x)$.

\subsection{Temperature Sampling Variants} 
There are many heuristics for selecting a short-list of most likely token candidates, and they rely on different statistics of the model distribution. \textbf{Top-k} always selects top $k$ most probable tokens \citep{fan-etal-2018-hierarchical}. 
\textbf{Top-p} \citep{Holtzman2020TheCuriousCase} selects top tokens whose cumulative probability exceeds a hyperparameter $p$. 
Recently, \citet{nguyen2024turningheatminpsampling} proposed \textbf{min-p} that uses the discounted probability of the top-$1$ token to define the probability threshold. 
\textbf{$\epsilon$-sampling} \citep{ hewitt-etal-2022-truncation} allows any token with a probability greater than $\epsilon$. 
Some of the methods rely on the Shannon entropy of the model distribution \citep{shannon1948mathematical}, among which \textbf{$\eta$-sampling} truncates words below an entropy-dependent probability threshold \citep{ hewitt-etal-2022-truncation}. \citet{zhang2024edtimprovinglargelanguage} propose \textbf{EDT} (Entropy-based Dynamic Temperature) sampling in which they dynamically adjust the temperature based on entropy. 
These methods generally lead to better results than sampling directly from the unmodified $p(v|x)$, and improve diversity and accuracy. 
In \Cref{sec:results}, we find that even the best among them still exhibit quality degradation when increasing the temperature. 
We provide a more detailed description of the aforementioned methods in \Cref{app:sampling_methods}.

%% file: sections/sampling_risk_analysis.tex
\section{When Do the Sampling Methods Fail?}
\label{sec:analysis}
In this section, we analyze the limitations of widely used sampling methods introduced in \Cref{sec:preliminaries} by focusing on their impact on specific decoding time steps. 
We aim to identify sensitive positions where sampling significantly increases the risk of generating incorrect outputs compared to greedy decoding. 
To quantify this, we propose a metric \textit{\risk} that measures the likelihood of output errors when replacing greedy tokens with sampled ones.
Our findings highlight the need for adaptive decoding strategies that switch selectively between greedy and sampling to maintain output quality.

\subsection{Sampling Risk}

\begin{figure}[t] 
    \centering
     \includegraphics[width=1.0\linewidth]{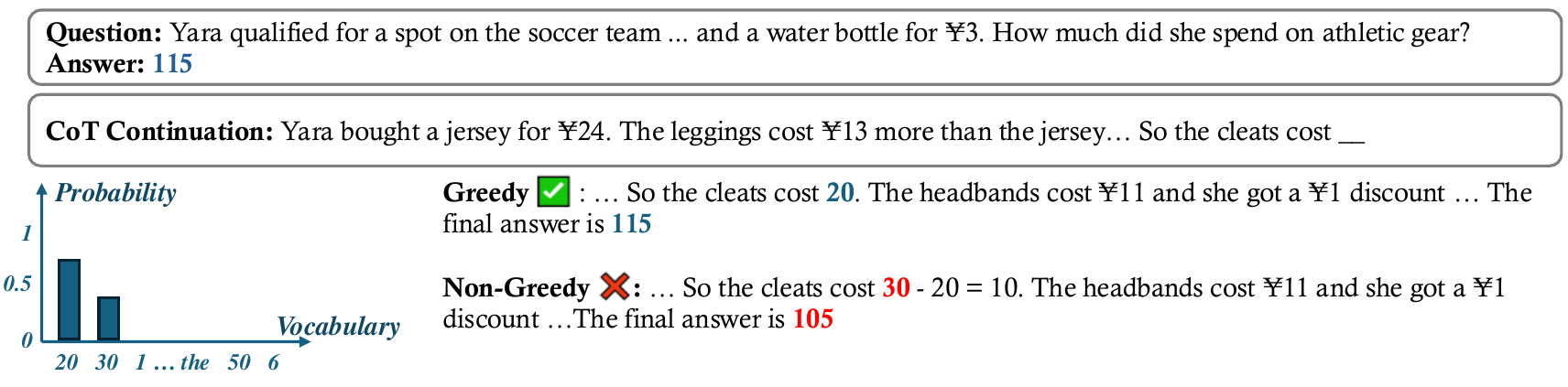}
    \caption{High \risk example for the current decoding position. \textbf{CoT Continuation} is generated by greedy decoding. \textbf{Greedy} continuation $20$ results in a correct answer while \textbf{non-greedy} continuation $30$ leads to an incorrect answer. 
    The full example is shown in \Cref{app:examples}, \Cref{case_studies}.}
    \label{fig:example_analysis}
\end{figure}

Motivated by the concept of \textit{regret} in reinforcement learning \citep{regret_bubeck}, we want to know the risk of choosing a sampling method in the current state compared to greedy decoding. 
We define \textbf{\risk} of a current prefix $x$ as follows:
\begin{equation}
\label{eq:risk}
\text{s-risk}(x) := R(x) - \mathbb{E}_{v\sim p} \left[ R\left([x,v]\right) \right],
\end{equation} where $[x,v]$ denotes the concatenation of the current prefix with a sampled next token, and $R(x)$ is the reward obtained by continuing from $x$ with only greedy tokens until the stopping criteria is met. 
In our case, we use accuracy as a reward \footnote{Accuracy is measured by matching between predicted and gold final answers.}. 
A higher \risk value indicates a greater likelihood of generating incorrect outputs when applying temperature sampling for this decoding time step.

\subsection{Case Studies of High Sampling Risk}
To validate the definition of the sampling risk, we first conduct a case study to investigate whether we can identify certain decoding time steps with high sampling risk. 
For this study, we focus on an arithmetic reasoning task. We hypothesize that when a model produces an arithmetic calculation, there are certain decoding steps with high \risk. 

\subsubsection{Setup} We focus on an arithmetic reasoning task with the chain-of-thought GSM-Symbolic dataset \citep{mirzadeh2025gsmsymbolic}, and use the instruction-tuned LLaMa-3.1-8B model \citep{grattafiori2024llama3herdmodels} as our language model. 
For this analysis, we subsampled 100 correct greedy outputs from the CoT GSM-Symbolic dataset to ensure our observations were not limited to a few cases. 
Within these outputs, we identified potentially risky token positions considering the positions where the model's top-$1$ token is an integer number. 
These positions are critical for math tasks, as integer numbers often serve as intermediate results for arriving at the final answer.

\begin{figure}[t]
    \centering
    \begin{subfigure}{0.32\textwidth}
        \centering
        \includegraphics[width=\textwidth]{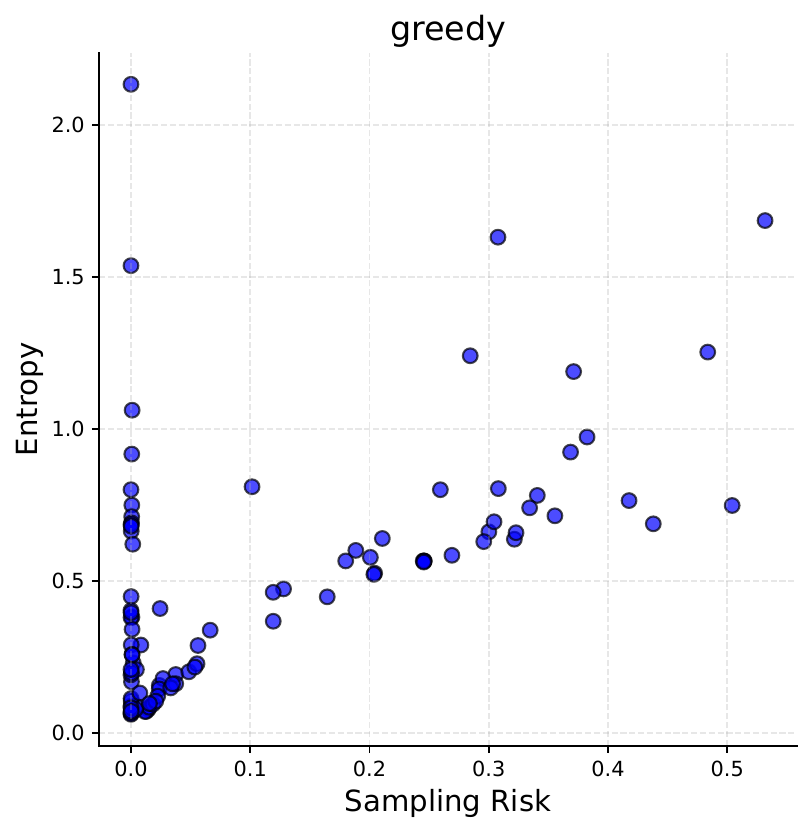}
        \caption{Greedy}\label{fig:greedy}
    \end{subfigure}
    \hfill
    \begin{subfigure}{0.32\textwidth}
        \centering
        \includegraphics[width=\textwidth]{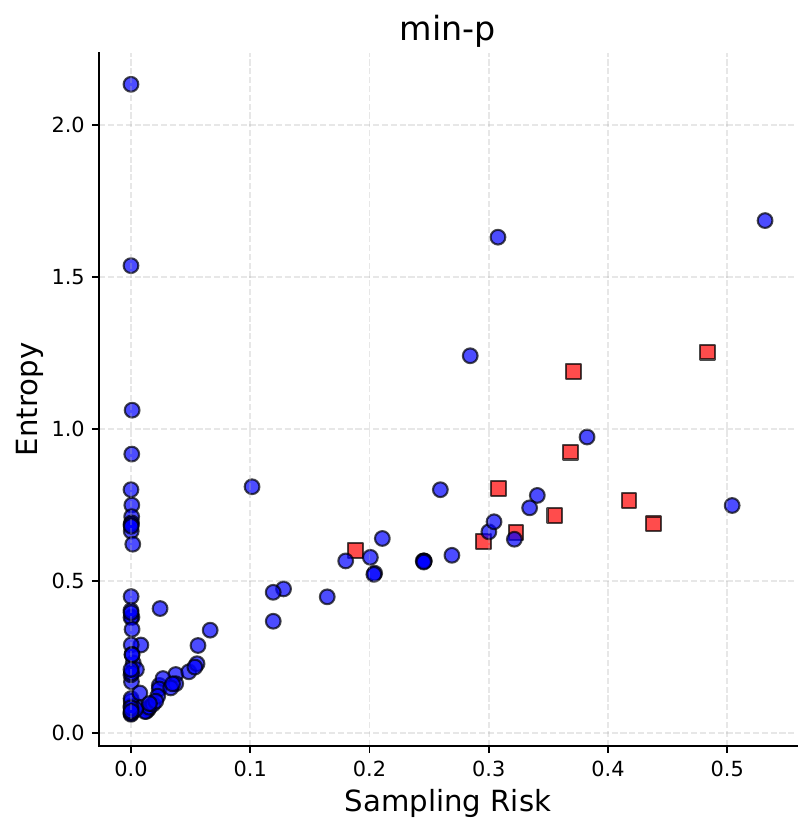}
        \caption{min-$p$ Sampling ($p=0.1$) }\label{fig:min-p}
    \end{subfigure}
    \hfill
    \begin{subfigure}{0.32\textwidth}
        \centering
        \includegraphics[width=\textwidth]{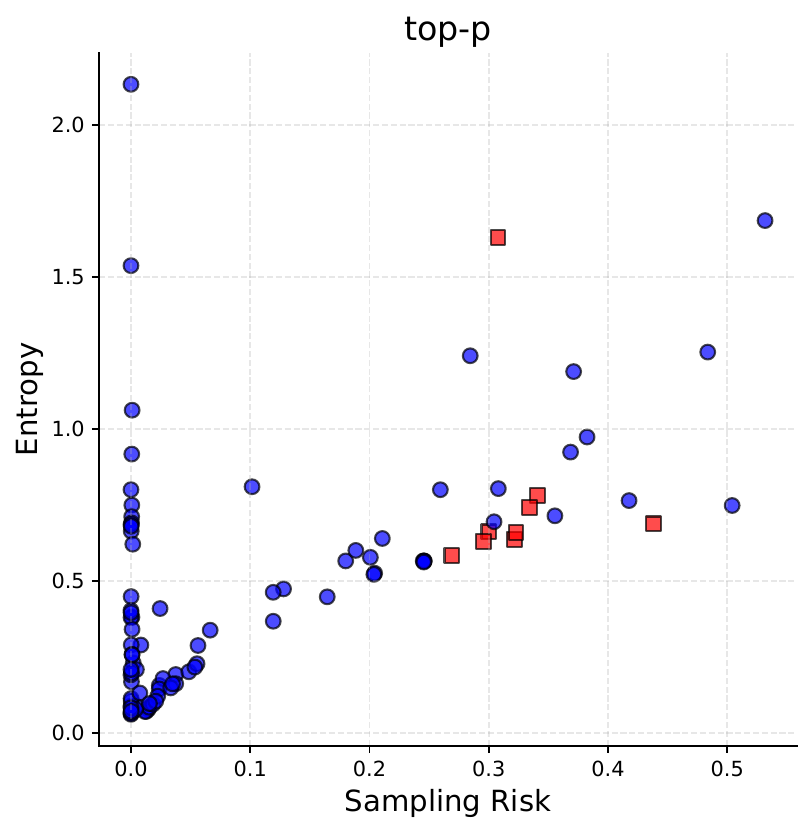}
        \caption{top-$p$ Sampling ($p=0.9$)}\label{fig:top-p}
    \end{subfigure}
    \caption{Entropy \textit{vs.} Sampling Risk. \textcolor{red}{Red} squares: Incorrect final answers; \textcolor{blue}{Blue} dots: Correct final answers. Overall, we show that introducing sampling methods for high-risk token positions will lead to erroneous outputs. }
    \label{fig:analysis_plots}
\end{figure}

\subsubsection{Findings}

\Cref{fig:example_analysis} illustrates one high \risk example from the GSM-Symbolic dataset. 
It shows that selecting a non-greedy token (e.g., $30$) for the current greedy CoT continuation leads to an incorrect answer. 
This suggests that sampling under the model distribution at the selected integer positions tends to generate errors, as only a narrow set of candidates can yield correct outputs. 

To further examine the impact of sampling at the selected positions, we compare min-$p$ and top-$p$ sampling with greedy decoding.
Figure~\ref{fig:analysis_plots} shows the correlation between \risk and entropy for different decoding settings.
We focus on full CoT continuations that produce correct answers with greedy decoding and compute \risk and entropy at integer positions within these continuations.
Overall, we show that there are certain decoding time steps when sampling tends to result in incorrect outputs, which also correlate with high model entropy. 
These results underscore the importance of the \risk metric in guiding decoding methods to improve the quality-diversity trade-off.

%% file: sections/methodology.tex
\section{Methodology: Selective Sampling}
\label{sec:method}
Motivated by the observation that high sampling risk leads to incorrect predictions (\Cref{sec:analysis}), in this section, we propose the \textbf{selective sampling} strategy by training a classifier $s_\theta(x) \in  \mathbb{R}_+$ to discriminate between safe and high-risk prefixes $x$.

\subsection{Estimating Sampling Risk}
We automatically label sampling risks defined in \Cref{eq:risk} for a small training set to train a classifier for selective sampling. 
For the training set, we either use a part of the training splits of the considered problems (GSM8k, Minerva Prealgebra tasks), or we split the test set into train and test (GSM-Symbolic task) in a proportion $60:40$. 
For the GSM-Symbolic task, we split the data using the original problem IDs, such that versions of the same problem only appear in one subset.

For a given task, we assume that a training example contains a source prompt $u$, and we can estimate \risk for a given decoding time step. 
We do not use any correct chain-of-thoughts from the data, and only use the model outputs. We only use correct greedy outputs for training.
We mark a prefix $x$ to be either risky or safe to sample from as follows:
\begin{equation}
    y(x)=\begin{cases}
        1, & \text{if s-risk}(x) < 1-\epsilon, \\
        0, & \text{otherwise}
    \end{cases},
\end{equation} where we set $\epsilon=0.05$ in our experiments.

In practice, we observe that for the reasoning tasks we considered, greedy sampling often produces high-quality outputs. Given a set of training prompts $U=\{u\}_{i=1}^{N}$, we obtain the subset of prompts and \textit{correct} greedy continuations $U_+=\{(u,x) | u \in U, R(u)=1\}$. Then, for every $(u,x) \in U_+$, we estimate sampling risks up to a maximal token position $M$ (we use at most $M=300$). To estimate the sampling risk at position $i$ for the current $x_{:i}$, we create a short-list of the top $k$ most probable next tokens. 
We force decode each of the $k$ next token candidates $\{v_j\}_{j=1}^{k}$, and then we finish each candidate using greedy decoding to estimate $R([x,v_j])$. Here, we treat greedy continuation as a low-cost approximation to the upperbound on the quality given a selected next token. Different ways to estimate this property may be possible and may depend on the task or the domain.

\subsection{Selective Sampling}
\begin{figure}
    \centering
    \includegraphics[width=1.0\linewidth]{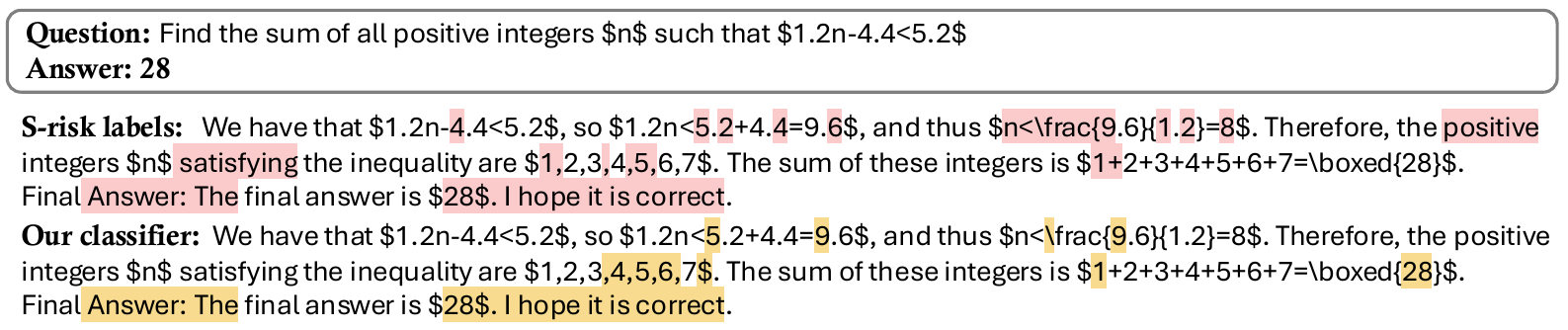}
    \caption{\textbf{S-risk labels}: obtained \risk labels for the Minerva task validation set; \textbf{Our classifier}: corresponding predictions from our classifier. Orange marks the high \risk positions where the greedy action would be chosen.}
    \label{fig:demo_classifier}
\end{figure}

Our goal is to train a classifier to predict the sampling risk based on a representation of the model's context. 
LLMs' hidden representations are known to be a rich source of information for feature extraction \citep{duan2024llmsknowhallucinationempirical, mahaut-etal-2024-factual}.
To obtain high-quality context features for the classifier, we simply use the hidden representations of the last token position. Namely, for a prefix $x=[x_1,\cdots,x_t]$, we use hidden states of the base model from the residual stream, including the last input embedding, for the last position $t$: $f(x)=[h_1, \cdots, h_L]$, where L is the number of layers in the base model. Then, the classifier is applied on top of the frozen hidden states $f(x)$. We parametrize a simple linear binary classifier $s_\theta(f(x))=\sigma(\frac{1}{L}\sum_{i=1}^L w_i^Th_i$), where $\sigma(y)=1/(1+e^{-y})$. 
Our classifier is a simple linear model applied on top of the hidden states, introducing minimal computational overhead. 
Most inference frameworks support access to the last hidden states, and we implement it directly inside the vLLM implementation of the LLaMa model \citep{kwon2023efficient}. During inference, we use our selective sampling classifier together with a truncation sampling (we use min-p, $p=0.1$). The aim is to investigate how our classifier can complement the limitations of truncation sampling (\cref{sec:analysis}).

We train our classifier in a teacher-forcing regime by passing the inputs to the model together with target sampling risks. We use binary cross-entropy loss to train the classifier. 
We discuss the training details in the \Cref{app:classifier_training}. 
Our implementation is released as open-source.%
\footnote{\url{https://github.com/serjtroshin/selective_sampling}}

%% file: sections/experiments.tex
\section{Experiments}
\label{sec:exps}

We conduct a comprehensive evaluation of selective sampling, comparing it to existing sampling methods across multiple tasks. 
Our experiments aim to show the effectiveness of selective sampling in improving both the quality and diversity of the model's outputs, particularly at higher temperatures.  
\subsection{Tasks}  
Following \citet{grattafiori2024llama3herdmodels, openai2024gpt4technicalreport, chen-etal-2024-minprompt}, we evaluate on commonly used mathematical reasoning benchmarks:  

\begin{itemize}  
    \item \textbf{GSM8K} \citep{cobbe2021gsm8k}: A grade school math problem-solving benchmark.
    \item \textbf{GSM-\textit{Symbolic}} \citep{mirzadeh2025gsmsymbolic}: An extended variant of GSM8K with symbolic templates, designed to provide a more reliable assessment of reasoning ability.
    \item \textbf{Minerva MATH} \citep{hendrycksmath2021}: A dataset of competition-level mathematical problems. We conduct experiments on the PreAlgebra subset. 
\end{itemize}  

We follow the standard prompting formats implemented in Evaluation Harness \citep{eval-harness}. For GSM8k and GSM Symbolic, we use the standard 8-shot chain-of-thought prompting configuration. For Minerva MATH, we use the 4-shot chain-of-thought prompting configuration. Additionally, we experiment on MMLU-Pro question answering task \citep{wang2024mmlupro} with multiple choice answers and CoT (\Cref{app:mmlu}).

\subsection{Model}  
We use the instruction-tuned LLaMa-3.1-Instruct (8B) model \citep{grattafiori2024llama3herdmodels}, selected for its strong performance across various reasoning tasks.

\subsection{Evaluation Metrics}
\paragraph{Quality-diversity trade-off evaluation.}
We evaluate the model's output based on two key aspects: \textbf{quality} and \textbf{diversity}, using $25$ samples per prompt.
Quality is measured by the accuracy of the final answer, indicating whether the model generates correct solutions.  
For diversity, we compute the averaged \textit{distinct $n$-grams} \citep{li-etal-2016-diversity}, which quantifies the proportion of unique $n$-grams relative to the total number of $n$-grams in the generated responses, details in \Cref{app:diversity_formula}. We follow \cite{nguyen2024turningheatminpsampling} and measure diversity only over correctly generated samples in order to be less biased towards low-quality outputs. 

To present results, we vary the temperature parameter and plot the quality metric on the $x$-axis and the diversity metric on the $y$-axis. The best method would have both high-quality and diverse predictions. 
To compare different methods, we look at the gap between the diversity-quality plots, e.g., which method is better in terms of quality given a certain diversity value and vice versa. 
In order to provide an aggregated metric integrating out the temperature parameter, we also report the area under the quality-diversity plot (\textbf{AUC}) using the trapezoidal rule \footnote{%
We take the theoretical $(0,0)$ point to belong to all curves even if it is not realized by any hyperparameters in the experiment.}
implemented in scikit-learn \citep{pedregosa2011scikit}.

\paragraph{Fluency evaluation.}
To demonstrate the effectiveness of our method in generating diverse yet coherent samples, we evaluate the noisiness of samples generated at high temperatures. 
We use perplexity as a proxy for noise, with higher perplexity indicating a higher likelihood of incoherent or nonsensical text, as suggested by previous studies \citep{zhang2024noise,ankner2024perplexed,marion2023less}. Perplexity scores were computed using the Llama-2-7B-chat-hf model\footnote{\href{https://huggingface.co/meta-llama/Llama-2-7b-chat-hf}{https://huggingface.co/meta-llama/Llama-2-7b-chat-hf}}.
For each task, we analyze a random subset of 100 instances, generating 25 samples per instance and calculating the average perplexity of the samples as the score for that instance. The overall perplexity score is then computed as the average of the per-instance scores.

%% file: sections/results.tex
\section{Results} \label{sec:results}

\subsection{Diversity-Quality Results}
We compare our selective sampling with baseline methods, including min-p sampling \citep{nguyen2024turningheatminpsampling}, top-p sampling,  top-k \citep{fan-etal-2018-hierarchical}, \citep{Holtzman2020TheCuriousCase}, $\epsilon$ sampling \citep{hewitt-etal-2022-truncation}, $\eta$ sampling \citep{hewitt-etal-2022-truncation} and EDT
\citet{zhang2024edtimprovinglargelanguage}.
The full plots are listed in \Cref{fig:ablation_gsm8k,fig:ablation_minerva} (\Cref{app:examples}), where we demonstrate that the baselines methods usually follow the same diversity-quality trajectory, and that there is no clear winner among the baseline methods. 
In Appendix \ref{sec:hyperparameter}, we analyze the effect of the hyperparameters for the baselines, and we choose min-$p$, with $p=0.1$, as the main baseline for comparison. 
In \Cref{tab:main_result}, we compare the cumulative aggregated diversity-quality scores (AUC metric) for the quality-diversity plots.

In \Cref{fig:combined_plots_main}, we demonstrate that the quality-diversity trade-off of min-p can be improved using selective sampling for all three tasks. 
When the temperature is lower $(\tau \leq 0.5)$, the performance gap between min-p and ours is small. 
However, our method performs better than min-p when the temperature value is increased.

Moreover, we estimate the average percentage of token positions where selective sampling chooses greedy over temperature sampling, as shown in \Cref{tab:statistics_greed}. 
We observe that selective sampling tends to choose greedy decoding more often under higher temperature values and harder tasks, such as Minerva.   
This observation is expected, as higher temperatures or more challenging tasks tend to degrade the quality. 
Using greedy decoding more frequently in such cases results in a more reliable strategy to maintain the output quality.

\subsection{Fluency Results}
\Cref{fig:combined_plots-perplexity} illustrates the average perplexity scores across different temperature settings for GSM8K, Symbolic GSM and Minerva Prealgebra. It also highlights the average accuracy of the samples at each temperature value.
As shown in the figure, our method consistently produces significantly less noisy and more accurate samples, even at high temperatures, compared to the min-$p$ sampling method.
Examples of samples generated by our method versus min-$p$ at $\tau=2.0$ are presented in \Cref{app:examples}. 

\input{figures/our_vs_min_p}

\input{figures/perplexity}
\input{tables/auc}

\subsection{Ablation Studies}
\paragraph {Classifier features.}
Above, we demonstrated that our classifier trained on top of the last hidden states can improve the quality-diversity trade-off. 
While our classifier uses more information from the context, compared to the baselines, it is built on top of the model's hidden representations.
To complement the main experiment and to check whether we can train the classifier solely using the current context as the feature, we introduce a simple n-gram-based model, a classifier trained over the last $n$ input embedding representations. We provide the architectural details in \Cref{app:n_gram}. 
For the Minerva evaluation set, the n-gram classifier obtains 79\% accuracy and 0.72\% ROC AUC, which is worse compared to the classifier trained on the hidden states (85\% accuracy, and 0.78 ROC AUC). 
In \Cref{fig:n_gram}, we present the results for selective sampling with the n-gram classifier, where we demonstrate that the n-gram classifier is slightly better compared to the min-$p$ baseline, in term of diversity-quality trade-off, for the smaller temperature values range $\tau \in (0;2]$. However, the n-gram classifier does not reach the performance of our hidden states classifier. We think it is easier for the model trained on top of the hidden states to generalize to the unseen contexts, especially in the higher temperature regime. 

\paragraph{Classifier sensitivity.}
To assess the sample complexity of training the classifier, in \Cref{app:sensitivity}
we compare performance with random subsamples of the training set.
We observe that performance convergens at around 500 training prompts,
suggesting that variations in training data do not pose problems to the framework.

\paragraph{Diversity metric choice.}
To verify that our results are not overdependent on the choice of the diversity metric, we follow the RFT \citep{yuan2023scaling} and use the diversity metric as the average normalized Levenshtein distance between all pairs of correct responses and compare our approach to min-p sampling. We find that our method improves the quality-diversity trade-off, in line with the findings from the n-gram-based diversity evaluation (see \Cref{fig:levenshtein} in \Cref{app:diversity_metric}).

\subsection{Entropy-based (EDT) baseline.}
In this experiment, we modify the EDT \citep{zhang2024edtimprovinglargelanguage} approach, where we use a binary threshold similar to our approach to switch between greedy and high-temperature settings. From the additional results (see \Cref{fig:edt_ablation} in \Cref{app:EDT_ablation_experiment}), we observe that the threshold-based entropy baseline does not outperform the entropy-based dynamic temperature sampling (EDT), and our method outperforms both variants of entropy-based sampling approaches, which highlights the benefit of the trained classifier head versus using entropy.

\subsection{Task Generalization of the Selective Sampling Classifier}
\label{sec:generalization_experiment}
\begin{figure}[t!]
    \centering
    \begin{minipage}{0.45\linewidth}
        \centering
        \includegraphics[width=\linewidth]{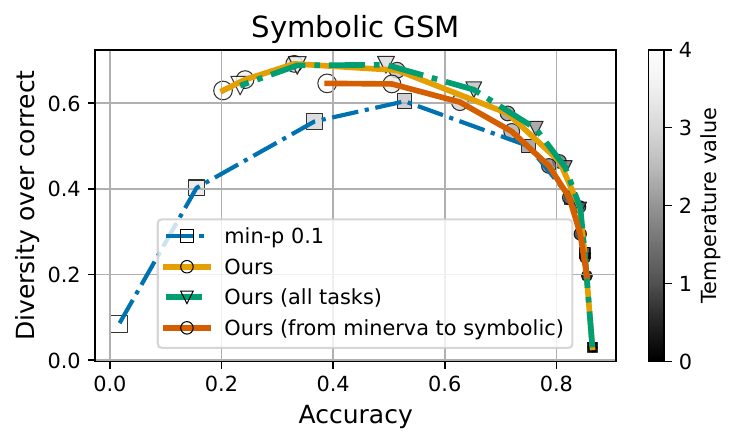}
    \end{minipage}
    \hfill
    \begin{minipage}{0.45\linewidth}
        \centering
        \includegraphics[width=\linewidth]{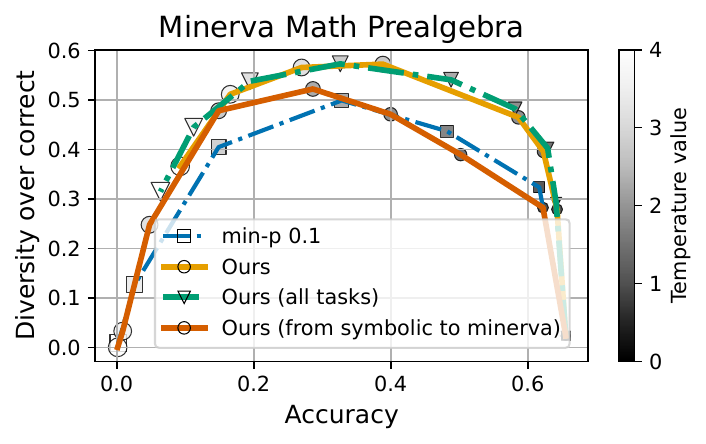}
    \end{minipage}
    \hfill
    \vspace{1em} %
    \caption{Generalization experiment with evaluation on Symbolic GSM and Minerva Math (Prealgebra) tasks. ``Ours (all tasks)'' is a classifier trained on all 3 datasets. ``Ours (from X to Y)'' denotes a transfer experiment, where we train the classifier on task X and test on Y.}
    \label{fig:generalization_2_tasks}
\end{figure}

In this experiment, we ask (1) whether we can transfer a classifier trained on one task to another task, (2) whether we can train a single classifier to be applied on many tasks. Regarding (1), we observe that selective sampling trained on the Minerva dataset outperforms the min-p baseline for the GSM Symbolic task on diversity-quality, suggesting that our classifier can generalize between these tasks (see \Cref{fig:generalization_2_tasks}). Regarding (2), we train a single classifier, Ours (all tasks), on 800 examples from each of the 3 datasets: GSM8k, GSM Symbolic, and Minerva, and evaluate the model on the GSM Symbolic task. We observe that the quality of the Ours (all tasks) model closely matches that of Ours trained only on GSM Symbolic (see \Cref{fig:generalization_2_tasks}, left). The same effect is observed when we evaluate the same all tasks model on the Minerva dataset (see \Cref{fig:generalization_2_tasks}, right). This shows that we can use the same single classifier on multiple tasks.

%% file: figures/our_vs_min_p.tex
\begin{figure}[t!]
    \centering
    \begin{subfigure}{0.305\linewidth}
        \centering
        \includegraphics[width=\linewidth]{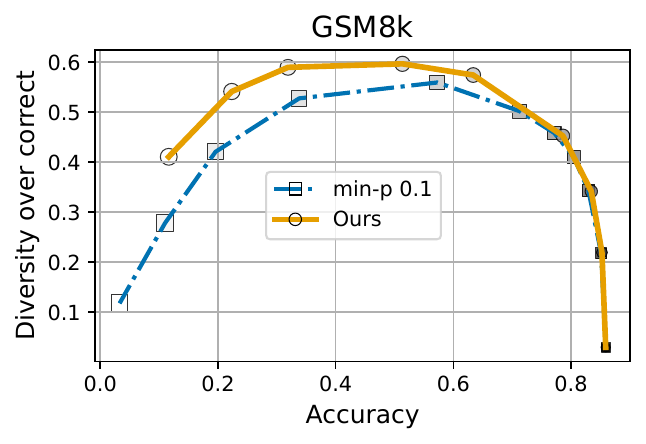}
        \caption{GSM8k}
    \end{subfigure}
    \hfill
    \begin{subfigure}{0.305\linewidth}
        \centering
        \includegraphics[width=\linewidth]{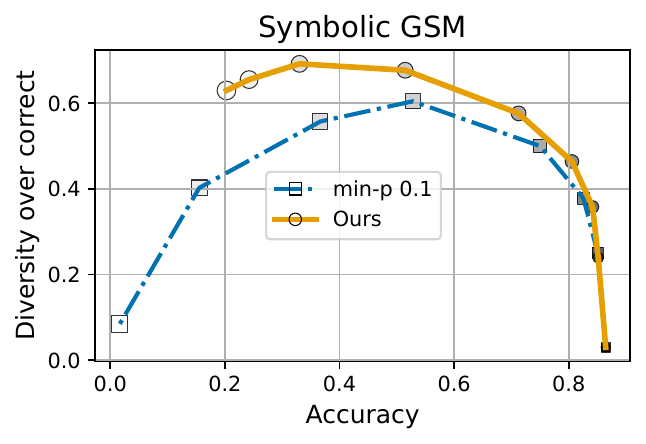}
        \caption{Symbolic GSM}
    \end{subfigure}
    \hfill
    \begin{subfigure}{0.333\linewidth}
        \centering
        \includegraphics[width=\linewidth]{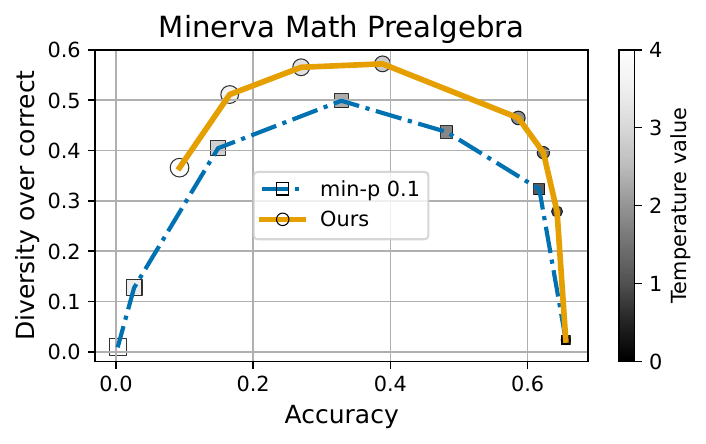}
        \caption{Minerva (Prealgebra)}
    \end{subfigure}
    \caption{Selective sampling with our classifier improves the diversity-quality trade-off compared to the strong min$-p$ truncation baseline. On the x-axis, we report the accuracy, and on x-axis, we report the Diversity over correct samples. Size and color of the circles mark the temperature parameter.}
    \label{fig:combined_plots_main}
\end{figure}

%% file: figures/perplexity.tex
\begin{figure}[t!]
    \centering
    \begin{subfigure}{0.31\linewidth}
        \centering
        \includegraphics[width=\linewidth]{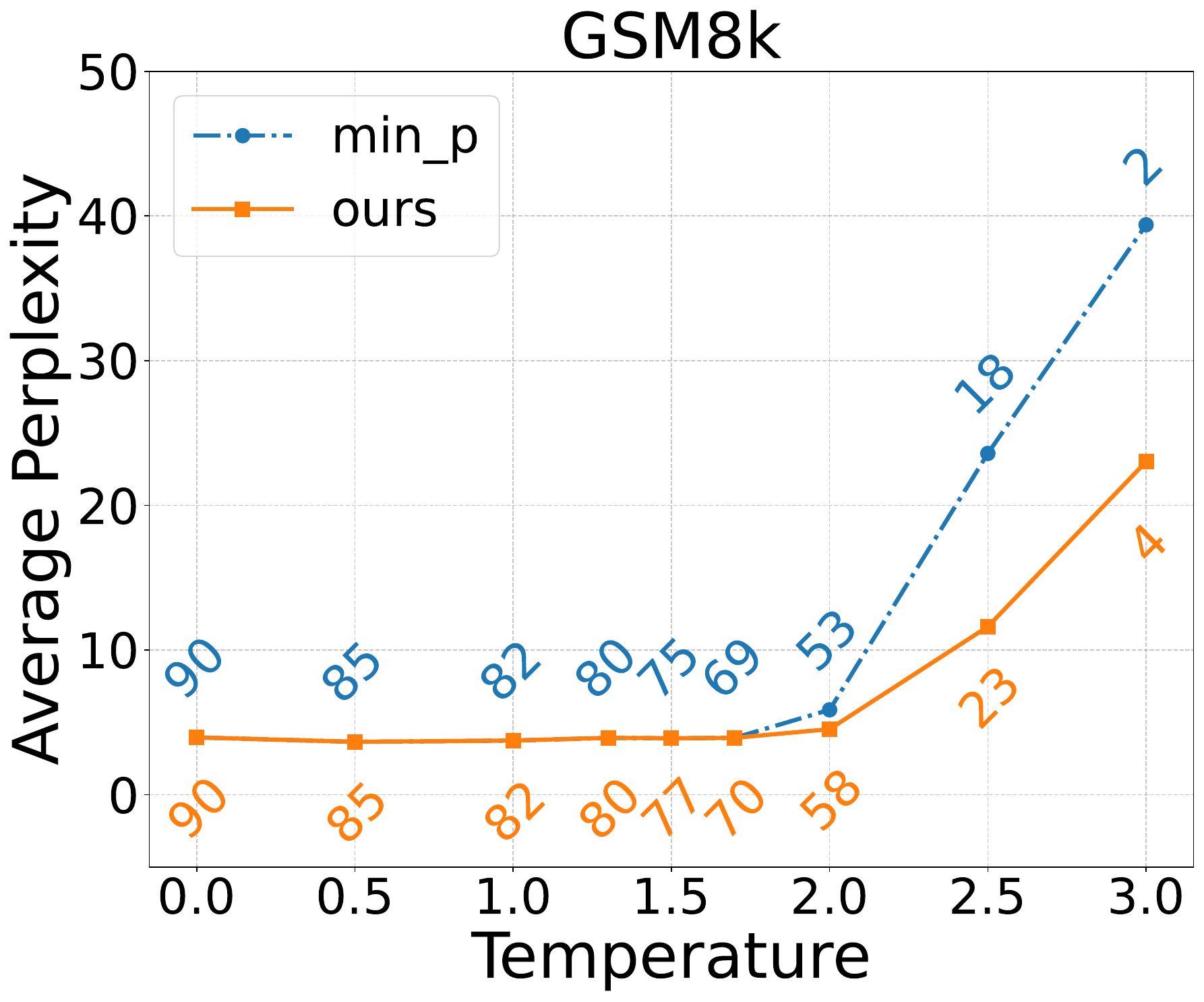}
        \caption{GSM8k}
        \label{fig:gsm8k-perplexity}
    \end{subfigure}
    \hfill
    \begin{subfigure}{0.31\linewidth}
        \centering
        \includegraphics[width=\linewidth]{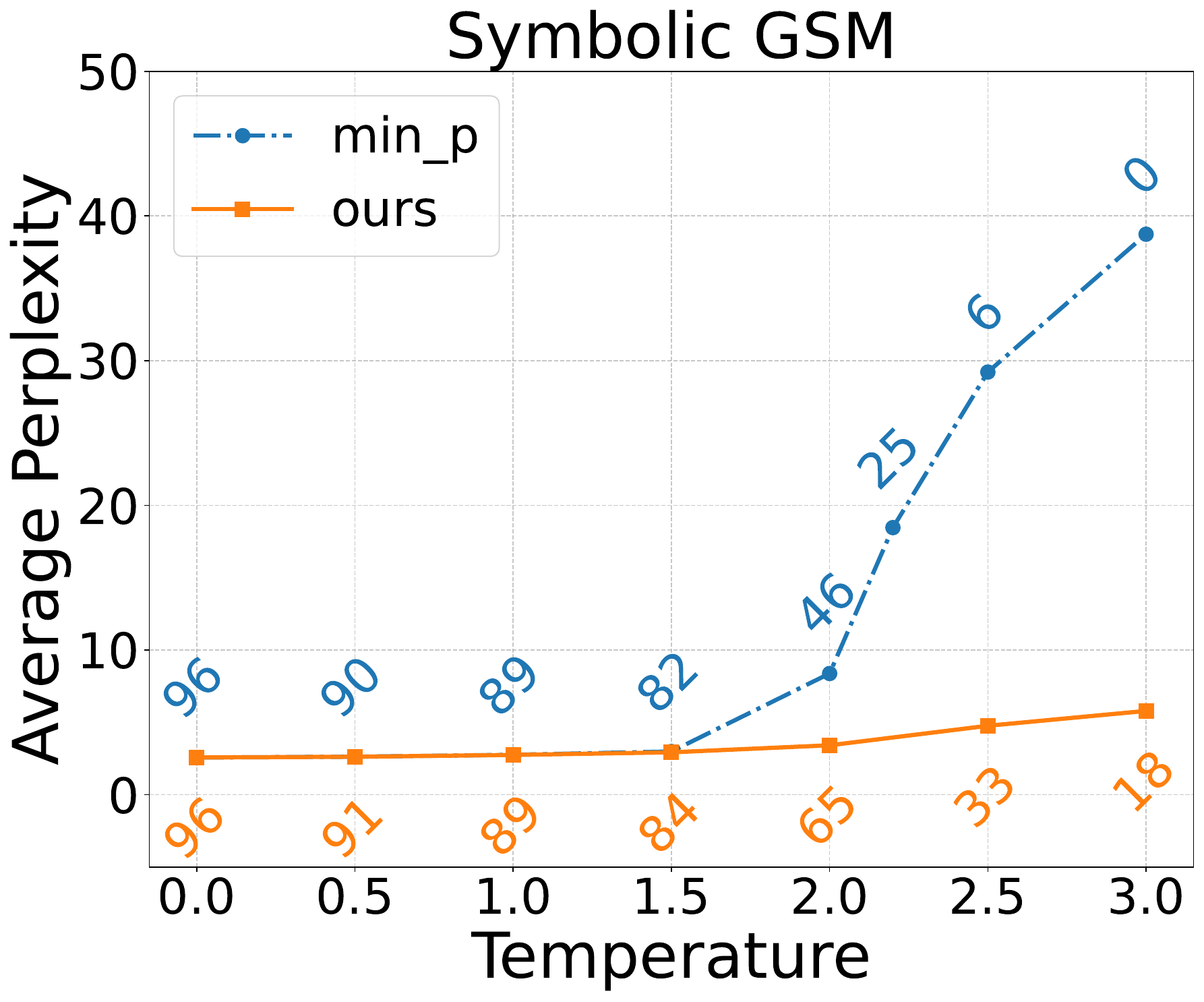}
        \caption{Symbolic GSM}
        \label{fig:symbolic-perplexity}
    \end{subfigure}
    \hfill
    \begin{subfigure}{0.31\linewidth}
        \centering
        \includegraphics[width=\linewidth]{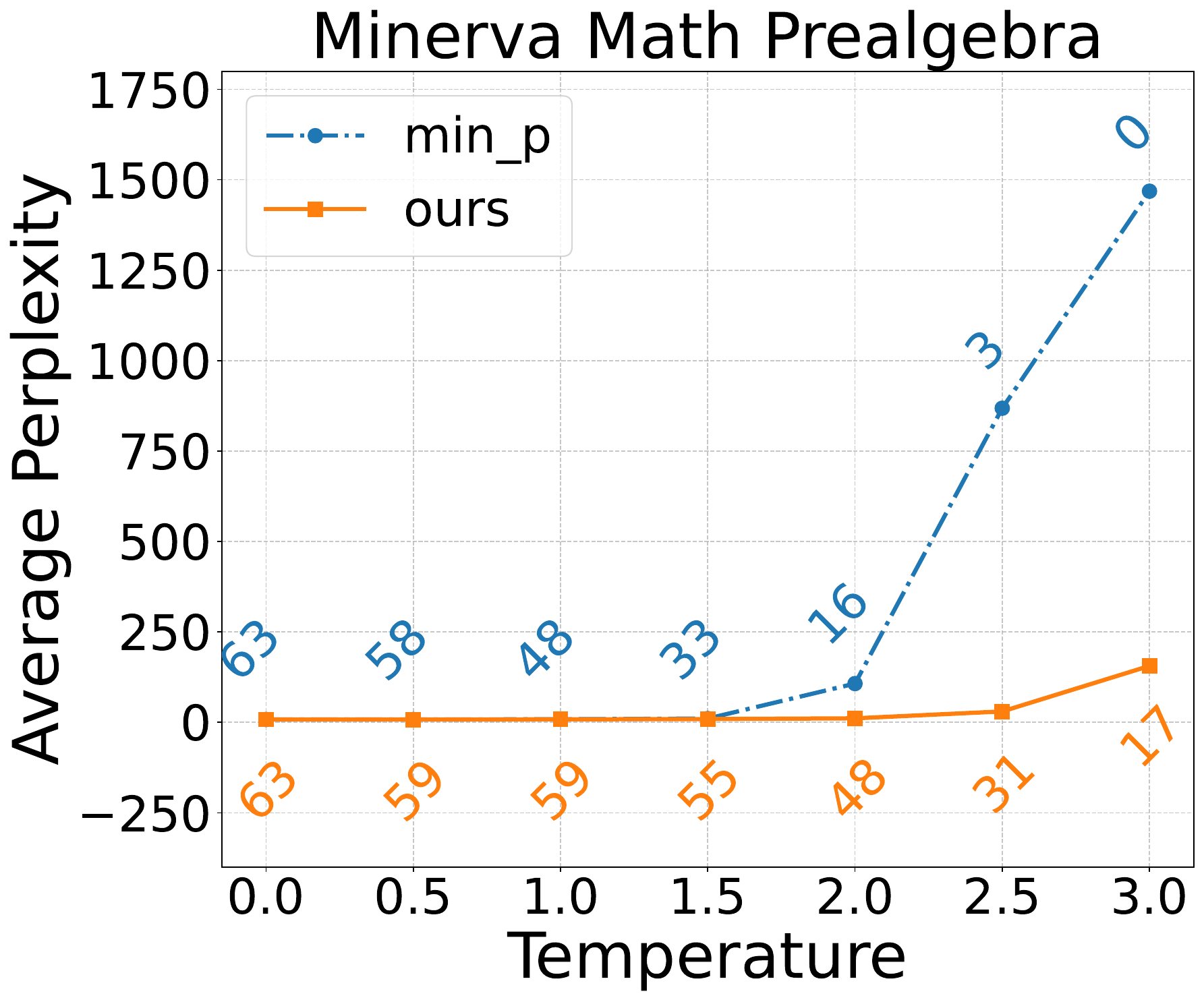}
        \caption{Minerva (Prealgebra)}
        \label{fig:minerva-perplexity}
    \end{subfigure}
    \caption{Our approach produces low perplexity samples and maintains better quality than the min-$p$ baseline even at high temperatures. Numbers inside the plots represent accuracies (as percentages).}
    \label{fig:combined_plots-perplexity}
\end{figure}

%% file: tables/auc.tex
\begin{figure}[h!]
    \centering
    \small
    \begin{minipage}{0.45\textwidth}
    \centering
    \setlength{\tabcolsep}{3pt}  %
    \begin{tabular}{l c c c}
        \toprule
         &  & Symbolic & Minerva \\ 
         Method                  &   GSM8k   &  GSM          & Prealgebra \\ 
        \midrule
        top-$p$             & 0.32    & 0.32 &  0.21   \\
        min-$p$             & 0.38    & 0.40 &  0.25    \\
        top-$k$             & 0.38    & 0.40 &  0.23   \\
        $\eta$ sampl.       & 0.37    & 0.40 &  0.24   \\
        $\epsilon$ sampl.   & 0.37    & 0.40 &  0.24   \\
        EDT                 & 0.35    &  0.36 & 0.24 \\
        Ours                & \textbf{0.42}     & \textbf{0.47}  &  \textbf{0.30}      \\
        \bottomrule
    \end{tabular}
    \setlength{\tabcolsep}{6pt}  %
    \captionof{table}{The area under the quality-diversity plot of various sampling strategies: an aggregated metric of the quality-diversity trade-off.
    Our method outperforms the baselines on this metric.}
    \label{tab:main_result}

    \end{minipage}
    \hfill
    \begin{minipage}{0.48\textwidth}
        \centering
        \includegraphics[width=0.95\linewidth]{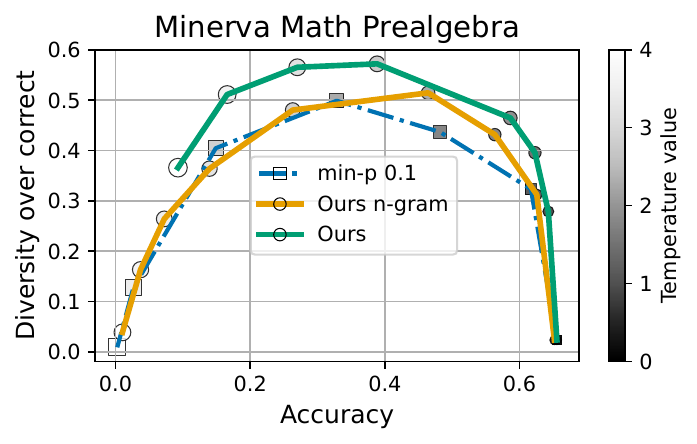}
        \caption{We show that our n-gram model, which does not use any internal model representations, can also slightly improve upon the min-p baseline.}
    \label{fig:n_gram}
    \end{minipage}
\end{figure}

 \begin{table}[h!]
     \centering
     \small
     \begin{tabular}{l l l l}
     \toprule
         Temperature & GSM8k & Symbolic & Minerva \\ 
         \midrule
        1.0           & 0.02    & 0.10 &  0.37   \\
        2.0       & 0.05    & 0.12 &  0.39    \\
        3.0        & 0.17    & 0.20 &  0.44   \\
        \bottomrule
     \end{tabular}
     \caption{Percentage of greedy decoding token positions under different temperatures and tasks for selective sampling. Under higher temperatures or harder tasks, selective sampling chooses greedy decoding more than high-temperature sampling.}
     \label{tab:statistics_greed}
\end{table}

%% file: sections/related_work.tex
\section{Related work}\label{sec:relwork}

In this section, we review prior work relevant to our study, focusing on sampling techniques from large language models (LLMs), methods for adjusting sampling parameters, and approaches to mitigating hallucinations. These areas collectively informed our approach for improving the quality and diversity of LLM outputs.

\paragraph{Sampling from LLMs.}  In the context of sampling from LLMs, 
various techniques \citep{Holtzman2020TheCuriousCase, basu2021mirostat, hewitt-etal-2022-truncation, nguyen2024turningheatminpsampling} are proposed to improve the fluency, coherence, reduction of repetitiveness, and diversity of LLM responses.
\citet{renze-2024-effect} finds that in context of problem solving through LLMs there is little difference between the existing sampling methods for $[0,1]$ temperature range. \citet{wang2024chainofthought} observe that LLMs can reason through sampling by considering alternative decoding paths. Minimum Bayes Risk (MBR) decoding \citep{kumar-byrne-2004-minimum} improves the quality of the model by countering the pathological behavior of MAP inference \citep{freitag-etal-2022-high, suzgun-etal-2023-follow, instruction-following-mbr, jinnai-etal-2024-generating}.

\paragraph{Adjusting LLM sampling.} There are several relevant approaches to improve sampling quality and creativity by adjusting the temperature of the model. \citet{liu-etal-2024-self-regulated} propose to dynamically choose sampling hyperparameters based on the context, where they use a prompted LLM to predict the hyperparameters at each step in a zero-shot regime. They assume LLMs already understand sampling risk. Their approach requires maintaining two models during the inference, which doubles the computation. \citet{chang2023kldivergenceguidedtemperaturesampling} develop a dynamic temperature sampling to improve the context grounding for question answering tasks, and \citet{zhang2024edtimprovinglargelanguage} improve upon this work by dynamically adjusting the temperature parameter based on the base model entropy. In our work, we use the approach of \citet{zhang2024edtimprovinglargelanguage} as an entropy-based baseline. In the context of code generation, \citet{zhu2023hotcoldadaptivetemperature} notice that challenging code tokens tend to appear at the beginning of a code block and propose to adjust the temperature based on the model confidence.

\paragraph{Mitigating LLM hallucinations.} A broader related line of work improves the quality of LLMs output by mitigating hallucination \citep{guerreiro-etal-2023-looking, duan2024llmsknowhallucinationempirical, mahaut-etal-2024-factual, chen2024hallucinationdetectionrobustlydiscerning}.
To detect hallucinations, \citet{duan2024llmsknowhallucinationempirical} and \citet{ mahaut-etal-2024-factual} use LLM hidden states to estimate its reliability.  
\citet{chen2024incontext} and \citet{meng-etal-2025-learn} adjust the next token prediction distribution to improve the factuality and overall quality by using the model entropy. 
\citet{Kossen2024SemanticEP} propose semantic entropy probes to approximate semantic entropy from the hidden states to mitigate hallucination. 
Moreover, \citet{li-etal-2023-contrastive} introduce extra models to guide the decoding while \citet{li2023inferencetime} shift the activation during the inference to improve the truthfulness of LLMs.

%% file: sections/conclusion.tex
\section{Conclusion}
\label{sec:conclusion}
Our work investigates temperature sampling from LLMs in the context of reasoning tasks. We highlight that fully relying on model confidence might lead to low-quality outputs. 
To complement the confidence estimate of a base model, we introduce the \textit{sampling risk} metric to estimate the risk of choosing a sampling action instead of greedy, based on the expected future task-specific reward. 
We then propose a selective sampling approach that switches between sampling and greedy decoding at each decoding time step based on the predicted sampling risk. 
To achieve this, we train a simple classifier to estimate the risk during inference.
We demonstrate that our selective sampling strategy can improve the quality-diversity trade-off compared to the commonly used baselines, such as min-$p$ truncation sampling. 
We hope our work will encourage further improvements on the important quality-diversity tradeoff in language modeling.

%% file: sections/app.tex
\section{Classifier Training}\label{app:classifier_training}
We train our classifier using Hugging Face\footnote{\url{https://huggingface.co/docs/transformers/en/main_classes/trainer}}. We train our classifier with the Adam optimizer \citep{kingma:adam} with $\beta_1=0.9, \beta_2=0.95, \epsilon=1\mathrm{e}{-12}$. We use learning rate $0.001$, weight decay $0.01$, and batch size $100$, number of epochs $50$. We use binary cross-entropy loss. We freeze all parameters of the base model and only train classifier weights.  To create the labelled dataset, we use only $k=8$ next token candidates to limit the complexity of the dataset creation, which is equivalent to using top-k filtered $p$  distribution for the regret estimation, and we use the temperature $\tau=3.0$) to estimate the sampling risk.

For preliminary experiments, we measure the quality of the binary classification with our hidden states classifier using the validation set ($100$ samples from the training set). For GSM8k/Minerva our classifier obtains 93\%/85\% accuracy, and 0.73/0.78 ROC AUC. We visualize the predictions of our model along with original labels in \Cref{fig:demo_classifier}. Overall, we observe that our classifier approximates well the labels in the dataset. We hypothesize that further data refinement and noise reduction can benefit classifier training. We report dataset statistics in the \Cref{tab:dataset-sizes} (for training, we leave only the samples with a correct greedy solution).

\begin{table}[ht]
\centering
\begin{tabular}{lrrr}
\toprule
Subset & GSM8k & GSM Symbolic & Minerva \\
\midrule
train  & 4487  & 2300     & 893     \\
val    & 100   & 100      & 100     \\
test   & 1319  & 2000     & 871     \\
\bottomrule
\end{tabular}
\caption{Dataset sizes for GSM8k, GSM Symbolic, and Minerva}
\label{tab:dataset-sizes}
\end{table}

\section{N-gram Classifier}\label{app:n_gram}
We implement the $n$-gram classifier as a feed-forward network applied on top of the last $n$ input embeddings (we chose $n=10$ based on the classification quality on the validation set). The network consists of the $1D$ Convolution layer with $m=256$ filters, which aggregates the last $n$ embeddings followed by the LayerNorm \citep{ba2016layernormalization}, and the GeLU nonlinearity \citep{hendrycks2023gaussianerrorlinearunits}. Then we apply an $m \times m$ linear layer following by the the LayerNorm and the GeLU, and finally followed by the last linear layer to project into single number.

\section{Sampling Methods}
\label{app:sampling_methods}
\subsection{Truncation Sampling Methods}
 Temperature samping modifies the logits by introducing the temperature parameter $\tau$: $z^\tau_{\text{LM}}(\cdot | x) = \tau \cdot z_{\text{LM}}(\cdot | x)$. Truncated temperature sampling as implemented in commonly used frameworks \citep[vLLM;][]{kwon2023efficient} works by 
 considering only short-list of token candidates $V' \subseteq V$ at each decoding step \citep{nguyen2024turningheatminpsampling, Holtzman2020TheCuriousCase, basu2021mirostat, hewitt-etal-2022-truncation}
 :
 \begin{equation}
 \label{eq:truncation}
    z(v|x) = \begin{cases}
      z^\tau_{\text{LM}}(v | x), & \text{if} \, v \in V', \\
      -\infty, & \text{otherwise}.
    \end{cases}
 \end{equation}
 and the next token is sampled from the categorical distribution:
\begin{equation}
\label{eq:Softmax}
    \tilde{p}^\tau(v|x) = \frac{\exp(z(v|x))}{\sum\limits_{v' \in V'} \exp( z(v'|x))}.
\end{equation}

\paragraph{Min-$p$ sampling.} To maintain coherence at high temperatures, \citet{nguyen2024turningheatminpsampling} introduced \textbf{min-$p$} sampling, a dynamic method that adapts its truncation threshold according to the model’s confidence at each decoding step. At each step, min-$p$ identifies the maximum probability token in the distribution:
\begin{equation}
    p_{max} = \text{max}_{v \in V} \, p(v|x).
\end{equation}
The truncation threshold is subsequently determined by scaling a base parameter $p \in (0,1]$ by $p_{max}$:
$p_{scaled} = p \, \times p_{max}$, and $V'$ (\cref{eq:truncation}) is determined by $V' = \{v \in V : p(v | x) \: \: \geq p_{scaled}\}$

\paragraph{Top-$p$ sampling.} Also known as nucleus sampling, this method samples from the "nucleus" of high-probability tokens by restricting the sampling pool to the top tokens whose cumulative
probability exceeds a hyperparameter $p$ \citep{Holtzman2020TheCuriousCase}: $V' = \{v \in V : \sum p(v | x) \: \: \geq p\}$

\paragraph{Top-$k$ sampling.} The sampling pool in this method consists of the most probable top $k$ tokens \citep{fan-etal-2018-hierarchical}: $V' = \{v \in V : \text{rank}\,(p(v | x)) \: \: \leq k\}$

\paragraph{$\epsilon$-sampling.} This method allows any token with a probability greater than a threshold $\epsilon$ \citep{ hewitt-etal-2022-truncation}: $V' = \{v \in V : p(v | x) \: \: > \epsilon\}$

\subsection{Entropy Sampling Methods}
For language models, Shannon entropy, defined as $H[p] = -\sum_{v \in V} p(v) \, \log(p(v))$ measures the uncertainty in predicting the next token in a sequence \citep{shannon1948mathematical}. 
Truncation sampling methods impose fixed thresholds to limit the set of candidate tokens during generation. While these methods can improve output quality by filtering out low-probability tokens, they may also over-restrict choices in low entropy distributions, potentially reducing diversity.
Entropy sampling methods adjust the sampling process based on the model's entropy, aiming to improve the quality-diversity trade-off. 

\paragraph{$\eta$-sampling.} $\eta$-sampling truncates words below an entropy-dependent probability threshold \citep{ hewitt-etal-2022-truncation}: $\eta = \min (\epsilon, \alpha \, \exp (-H[p]))$, using $\alpha \in [0,1]$ and the hyperparameter $\epsilon$. Token shortlist is then determined as $V' = \{v \in V : p(v | x) > \eta\}$.

\paragraph{EDT sampling.} Entropy-based Dynamic Temperature sampling method dynamically adjusts the temperature parameter according to the model's entropy and can be combined with truncation sampling methods as an initial step
\citep{zhang2024edtimprovinglargelanguage}. With the recommended hyperparameters  $\alpha=0.8$ and $\theta \in [0,1]$, they shrink the original temperature $\tau$ by  $f(H[p])$, namely $ T(\tau, H[p])= \tau \, \alpha^{\frac{\theta}{H[p]}}$. Intuitively, when the entropy is large, the shrinkage factor $f(H[p])$ approaches $1$, and when the entropy is small, it approaches $0$.

\section{Sampling Hyperparameter Settings}\label{sec:hyperparameter}

For comparing sampling methods: top-$k$, top-$p$, min-$p$, $\eta$/$\epsilon$, and entropy-based dynamic sampling (EDT), we report their optimal results with conducted tests on different hyperparameters in \Cref{tab:hyperparameter-ablation}. 
Results are selected based on the AUC scores over the quality-diversity plot, reflecting the overall quality-diversity trade-off.

For top-$p$ sampling, we report top-$p = 0.7$ and conducted tests on $p = 0.7, 0.8, 0.9$. 
For min-$p$ sampling, we report min-$p = 0.1$ for all experiments and conducted tests on $p = 0.1, 0.2, 0.3$. 
For $\eta$ and $\epsilon$ sampling, we test $\epsilon$ and $\eta$ values $0.0002$ and $0.0009$, found $0.0009$ to score better for both values and report this in our experiment results. 

For EDT sampling, we report $\theta = 0.1$ for all experiments, and conducted tests on $\theta = 0.1, 1.0$. 

\section{Averaged Distinct N-gram Diversity}\label{app:diversity_formula}
Given a set of test instances $S=\{s\}_{i=1}^{M}$, each with a response set $R_K = \{r\}_{i=1}^{K}$. Following \cite{nguyen2024turningheatminpsampling}, we first filter out incorrect samples for each instance in order to be less biased towards low-quality outputs, and we end up with a set of $C$ correct samples $R_C = \{r\}_{i=1}^{C}$ per instance, where $C$ may vary across instances. Then, for $n \in \{1,2,3,4,5\}$, we calculate the averaged distinct n-gram diversity for each instance as follows: 
\[\text{averaged distinct n-gram} \, (s) = \sum_{n=1}^{5} \frac{ \text{set}(\text{n-gram}(R_C))}{\text{n-gram}(R_C)}.\]
The total test-set diversity is the average diversity over all instances:
\[\text{diversity}(R_C) = \sum_{i=1}^{M} \text{averaged distinct n-gram} \, (s_i). \]

\section{Additional Experiments}
\subsection{MMLU (Social Tasks)}\label{app:mmlu}

\begin{figure}[t!]
    \centering
    \begin{minipage}{0.45\linewidth}
        \centering
        \includegraphics[width=\linewidth]{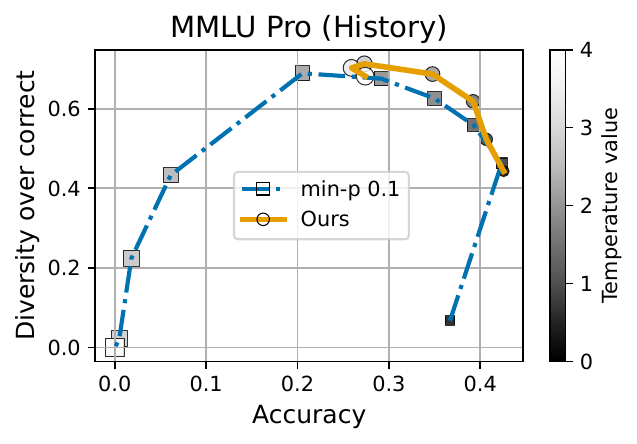}
    \end{minipage}
    \hfill
    \begin{minipage}{0.45\linewidth}
        \centering
        \includegraphics[width=\linewidth]{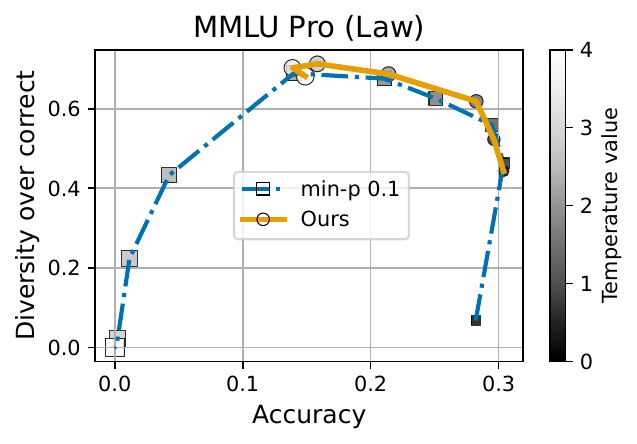}
    \end{minipage}
    \hfill
    \vspace{1em} %
    
    \centering
    \begin{minipage}{0.45\linewidth}
        \centering
        \includegraphics[width=\linewidth]{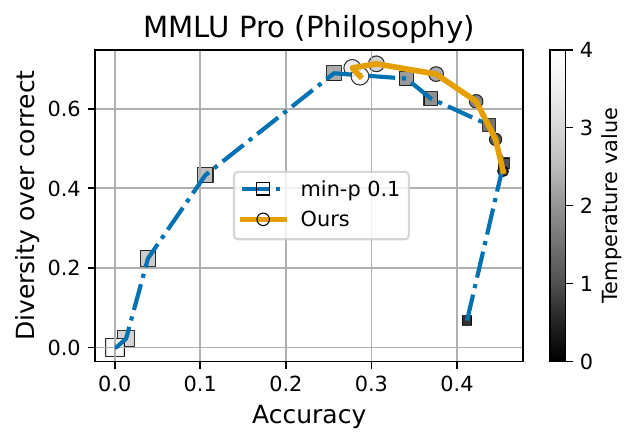}
    \end{minipage}
    \hfill
    \begin{minipage}{0.45\linewidth}
        \centering
        \includegraphics[width=\linewidth]{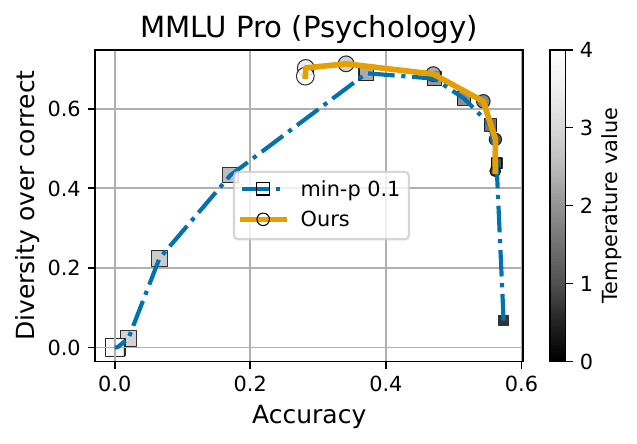}
    \end{minipage}
    \hfill
    \vspace{1em} %
    \caption{Additional result on MMLU Pro task on 4 social subsets. }
    \label{fig:mmlu_4_subtasks}
\end{figure}

\begin{figure}[t!]
    \centering
    \includegraphics[width=0.6\linewidth]{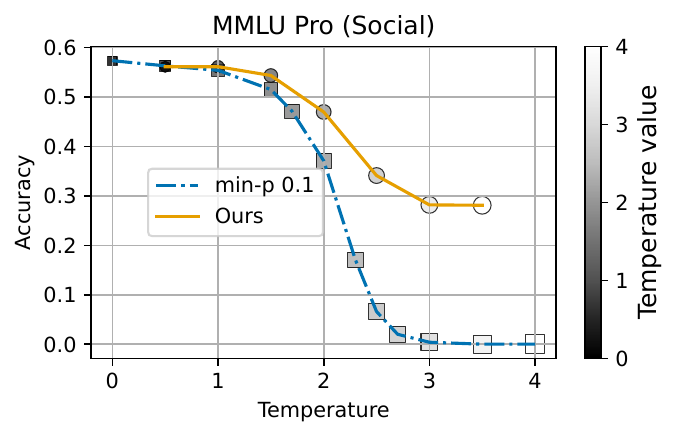}
    \caption{Additional result on MMLU Pro task on 4 social subsets: quality-temperature trade-off. }
    \label{fig:mmlu_temp_quality}
\end{figure}

We perform an additional experiment on the MMLU-Pro question answering task \citep{wang2024mmlupro} with multiple choice answers and CoT. We choose the subset of following tasks: law, philosophy, history, psychology, which are more different from the math tasks from our main experiments. As we can see from the results in \Cref{fig:mmlu_4_subtasks}, and \Cref{fig:mmlu_temp_quality}, (a) our method slightly outperforms the min-p baseline in terms of diversity-quality trade-off and (b) better preserves the quality for higher temperatures, which broadens the potential scope for our method.

\subsection{Sensitivity to the training dataset}\label{app:sensitivity}
\begin{figure}[t!]
    \centering
    \includegraphics[width=0.6\linewidth]{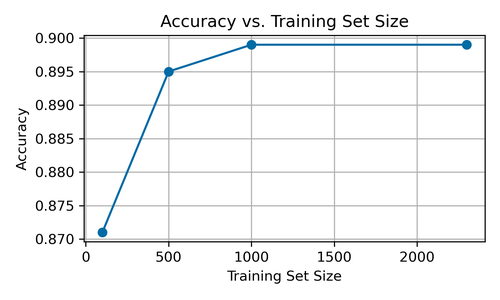}
    \caption{Training data sensitivity analysis on the GSM Symbolic task: accuracy vs the size of the training set. }
    \label{fig:sensitivity}
\end{figure}

To verify that our classifier training is not too sensitive for the variations in training data, we perform an additional experiment for classifier training by randomly subsampling 100, 500, 1000 examples of the training set for the GSM Symbolic task. We measure the accuracy for sampling risk classification on the validation set. From \Cref{fig:sensitivity}, we observe that subsampling 1000 or 500 examples reduces accuracy marginally from 0.9 to 0.89. Taking 100 examples leads to a slight overfitting and 0.87 accuracy.

\subsection{Ablation of the diversity metric}\label{app:diversity_metric}
\begin{figure}[t!]
    \centering
    \includegraphics[width=0.8\linewidth]{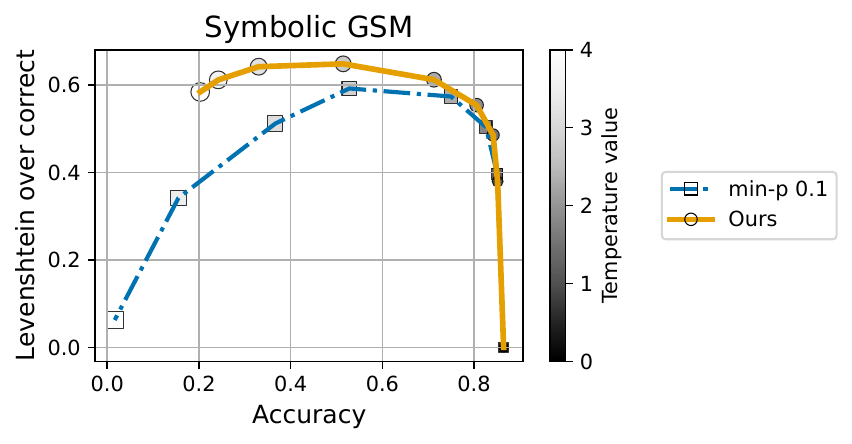}
    \caption{Quality-diversity trade-off plot with Levenshtein diversity metric. }
    \label{fig:levenshtein}
\end{figure}

Following RFT \citep{yuan2023scaling}, we use a diversity metric as the average normalized Levenshtein distance between all pairs of correct responses and compare our approach to min-p sampling. We find that our method improves the quality-diversity trade-off, in line with the findings from the n-gram-based diversity evaluation (see \Cref{fig:levenshtein}).

\subsection{EDT ablation experiment}\label{app:EDT_ablation_experiment}
\begin{figure}[t!]
    \centering
    \includegraphics[width=0.8\linewidth]{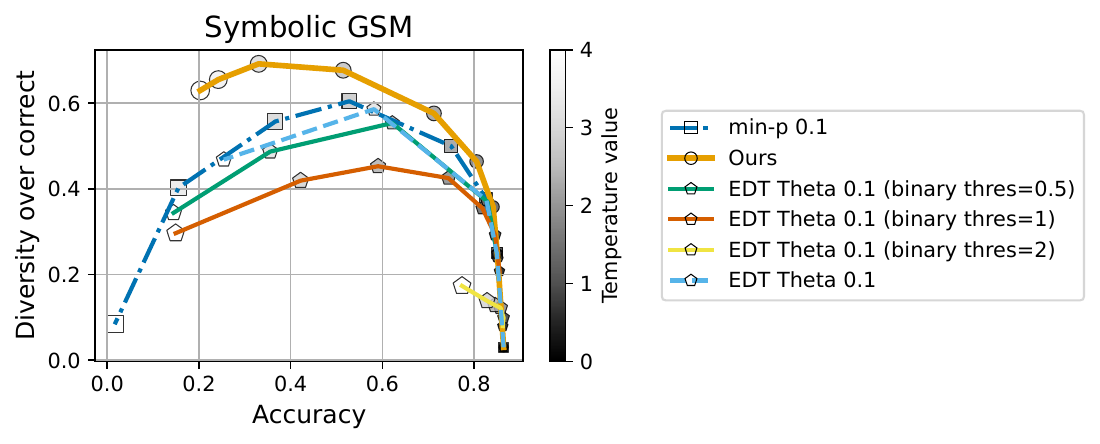}
    \caption{Quality-diversity trade-off plot with Levenshtein diversity metric. }
    \label{fig:edt_ablation}
\end{figure}

In this section, we perform an additional ablation experiment for the modified the EDT \citep{zhang2024edtimprovinglargelanguage} approach, where we use a binary threshold similar to our approach to switch between greedy and high-temperature settings. In particular, given a threshold $t$, we evaluate the threshold-based entropy baseline (if $H[p]$ < $t$, then $\tau'$=0.0, else $\tau'$=$\tau$), using $t$. from [0.5, 1, 2]. From the additional results \Cref{fig:edt_ablation}, we observe that the threshold-based entropy baseline does not outperform the entropy-based dynamic temperature sampling (EDT). Our method outperforms both variants of entropy-based sampling approaches, which highlights the benefit of the trained classifier head versus the entropy.

\section{Examples}
\label{app:examples}
\begin{figure}[ht] 
    \centering
    \begin{subfigure}{\linewidth}
        \centering
        \includegraphics[width=1.0\linewidth]{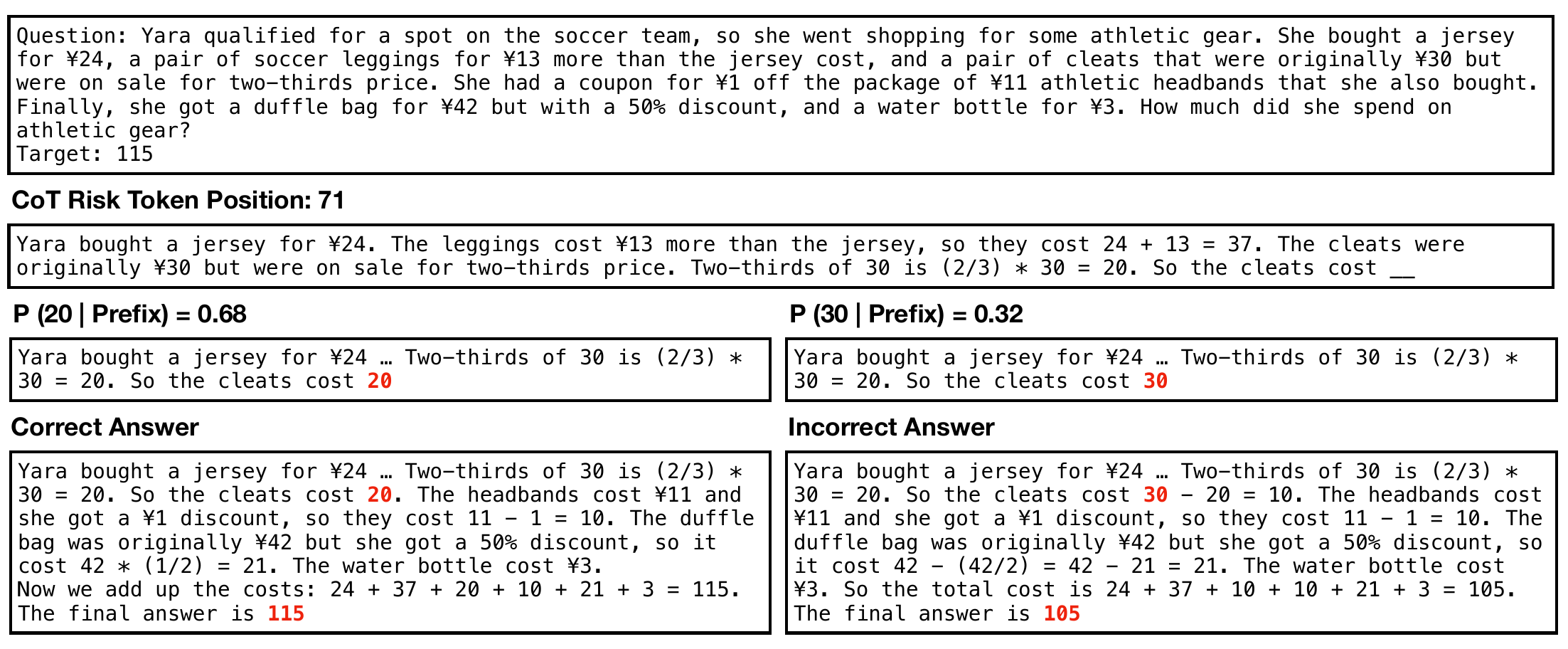}
        \caption{ Example 1 }
    \label{fig:analysis_sample_1}
    \end{subfigure}
    \vspace{1em} %
    \begin{subfigure}{\linewidth}
        \centering
        \includegraphics[width=1.0\linewidth]{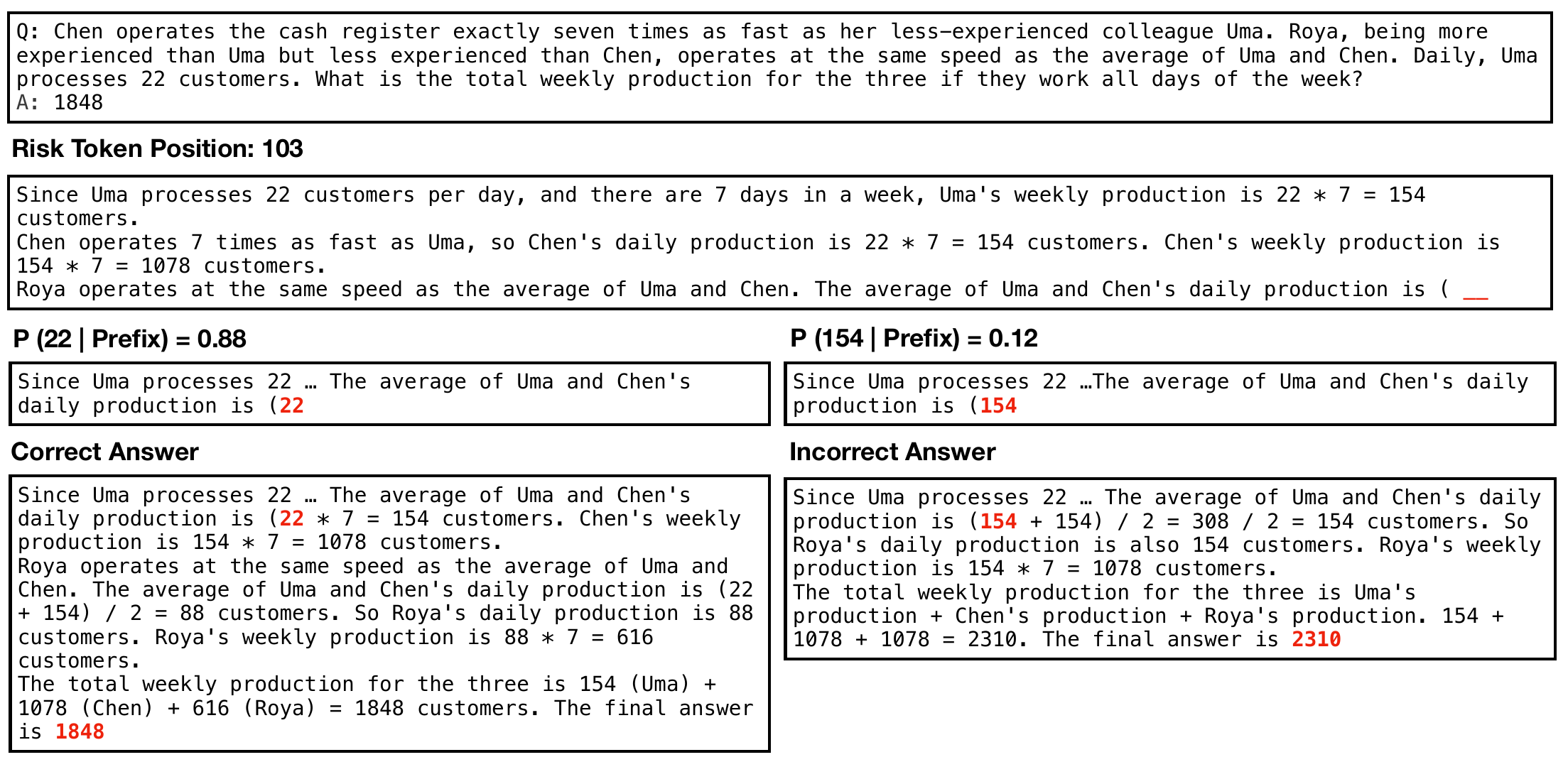}
        \caption{Example 2}
    \label{fig:analysis_sample_2}
    \end{subfigure}
    \caption{High-Risk sampling examples from GSM-Symbolic. \textbf{Left}: Top-1/Greedy generation on high-risk decoding position. \textbf{Right}: Non-greedy generation on high-risk decoding position.} \label{case_studies}
\end{figure}

\begin{figure}[ht]
    \centering
    \centering
    \includegraphics[width=1.0\linewidth]{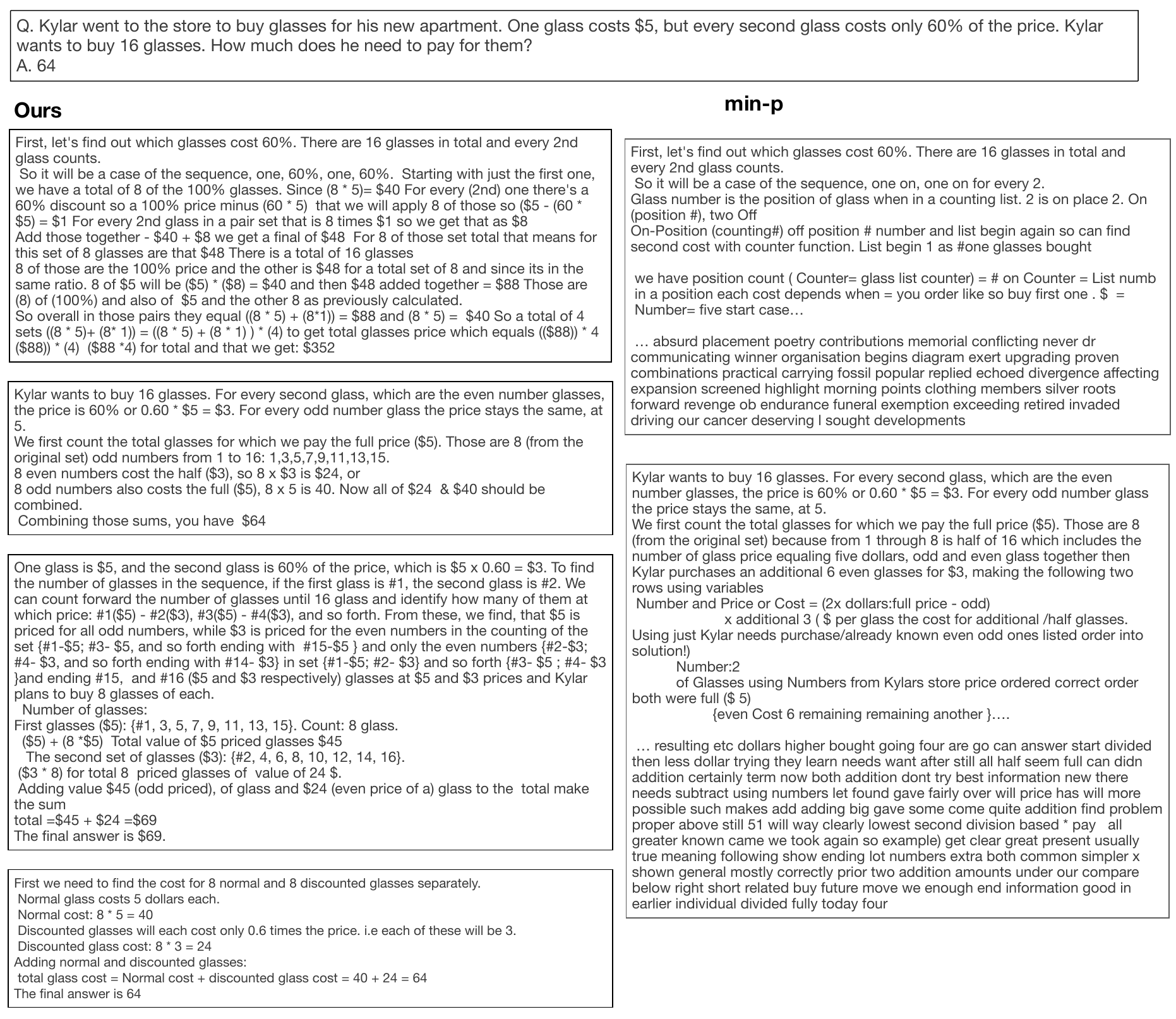}
    \caption{Example from GSM8k for Ours with temperature 2.0 versus min-p with temperature 2.0. Our model is more stable with higher temperatures compared to uncontrolled min-p, which fails co produce coherent outputs for some prompts with temperature $2$. We skip part of the output for min-p since it is too long.}
\label{fig:example_gsm8k}
    
\end{figure}

In \Cref{case_studies}, we provide a high sampling risk example from the Symbolic GSM task. In \Cref{fig:example_gsm8k}, we provide additional examples with $\tau=2.0$ for the samples for GSM8k. Overall, we observe that for higher temperatures, our methods produces more fluent outputs.

\begin{table}[h!]
\centering
\begin{tabular}{c|cc}
\toprule
Method & \multicolumn{2}{c}{Diversity-Quality \textbf{AUC}}  \\
& \textbf{GSM8K} & \textbf{Minerva} \\
\midrule
top-k 20  & 0.38  & 0.23 \\
\midrule
top-p 0.7 & 0.28 & 0.20   \\
top-p 0.8 & 0.31 & 0.20    \\
\textbf{top-p 0.9} & 0.32 & 0.21  \\
\midrule
\textbf{min-p 0.1} & 0.38 & 0.25 \\
min-p 0.2 & 0.37 & 0.25 \\
min-p 0.3 & 0.37 & 0.25 \\
\midrule
$\epsilon$-sampl. $0.0002$  & 0.35  & 0.23 \\
\textbf{$\epsilon$-sampl. $0.0009$}  & 0.37  & 0.24 \\
\midrule
$\eta$-sampl. $0.0002$  & 0.36 & 0.23 \\
\textbf{$\eta$-sampl. $0.0009$} & 0.37 & 0.24 \\
\midrule
\textbf{EDT $\theta\!=\!0.1$} & 0.35 & 0.24 \\
EDT $\theta\!=\!1.0$ & 0.32 & 0.24 \\
\bottomrule

\end{tabular}
\caption{Baselines hyperparameter values. Diversity-quality AUC Scores for GSM8K and Minerva Prealgebra. Bold denotes the chosen hyperparameter value for main results.}
\label{tab:hyperparameter-ablation}
\end{table}

\begin{figure}[t!]
    \centering
    \begin{minipage}{0.32\linewidth}
        \centering
        \includegraphics[width=\linewidth]{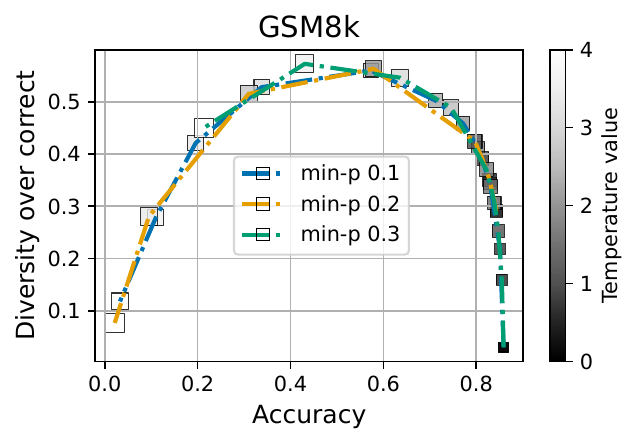}
        \caption*{GSM8k Min-p}
    \end{minipage}
    \hfill
    \begin{minipage}{0.32\linewidth}
        \centering
        \includegraphics[width=\linewidth]{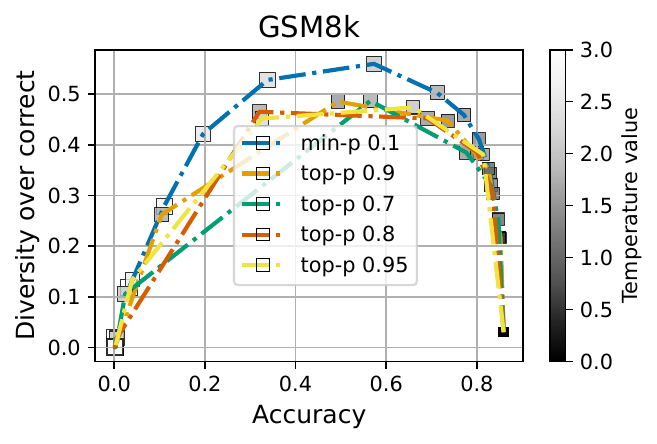}
        \caption*{GSM8k Top-p}
    \end{minipage}
    \hfill
    \begin{minipage}{0.32\linewidth}
        \centering
        \includegraphics[width=\linewidth]{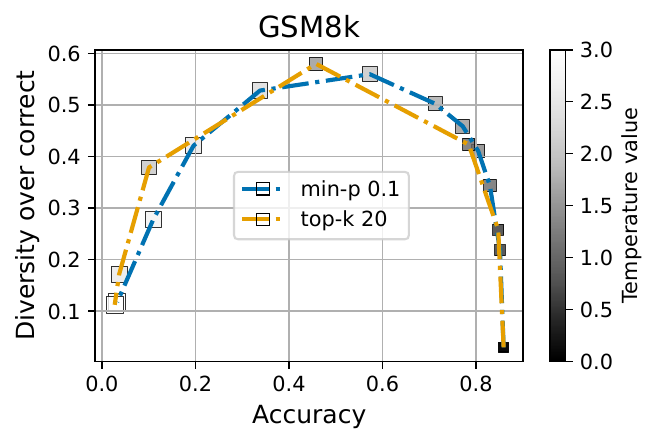}
        \caption*{GSM8k Top-k}
    \end{minipage}

    \vspace{1em} %
    
    \centering
    \begin{minipage}{0.32\linewidth}
        \centering
        \includegraphics[width=\linewidth]{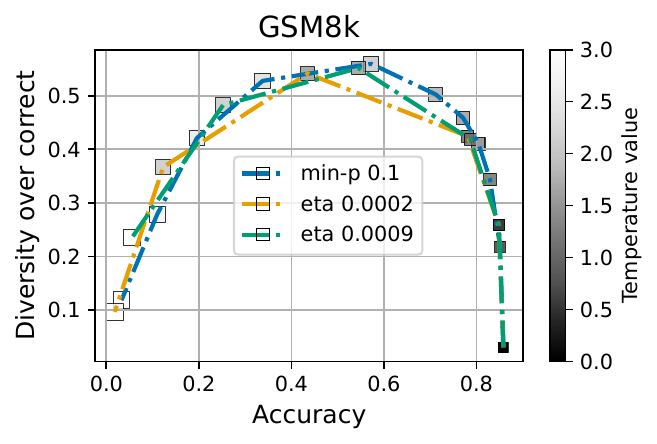}
        \caption*{GSM8k $\eta-$sampl.}
    \end{minipage}
    \hfill
    \begin{minipage}{0.32\linewidth}
        \centering
        \includegraphics[width=\linewidth]{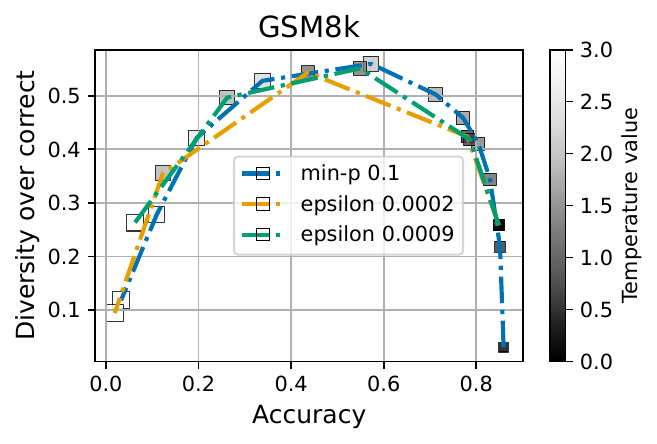}
        \caption*{GSM8k $\epsilon-$sampl.}
    \end{minipage}
    \hfill
    \begin{minipage}{0.32\linewidth}
        \centering
        \includegraphics[width=\linewidth]{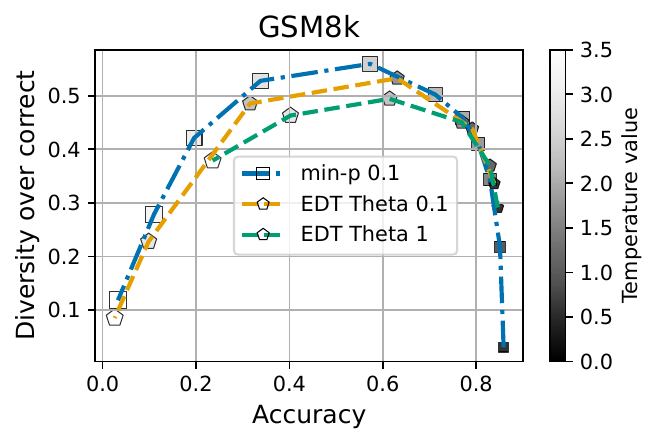}
        \caption*{GSM8k EDT}
    \end{minipage}
    \caption{We compare different hyperparameters for the baseline methods on the GSM8k task.}
    \label{fig:ablation_gsm8k}
\end{figure}

\begin{figure}[t!]
    \centering
    \begin{minipage}{0.32\linewidth}
        \centering
        \includegraphics[width=\linewidth]{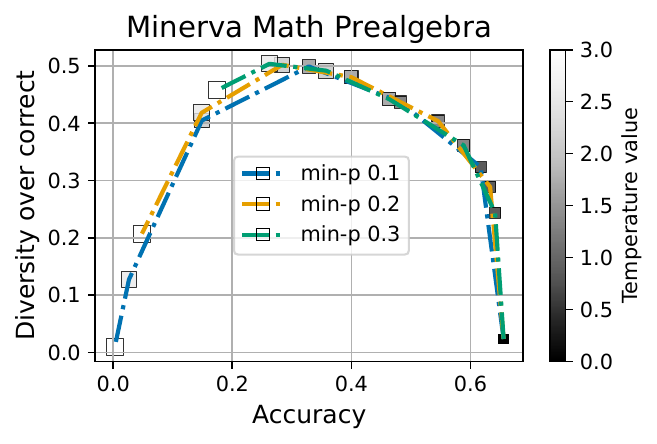}
        \caption*{Minerva Min$-p$}
    \end{minipage}
    \hfill
    \begin{minipage}{0.32\linewidth}
        \centering
        \includegraphics[width=\linewidth]{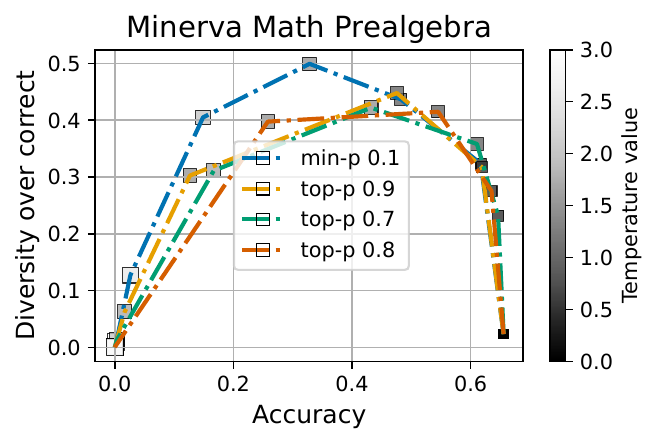}
        \caption*{Minerva Top$-p$}
    \end{minipage}
    \hfill
    \begin{minipage}{0.32\linewidth}
        \centering
        \includegraphics[width=\linewidth]{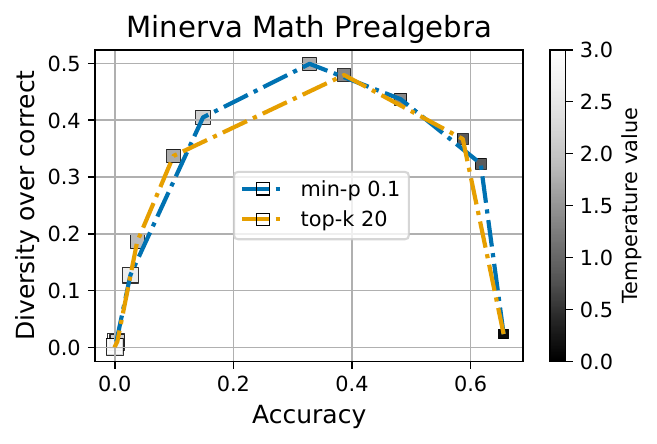}
        \caption*{Minerva Top-$k$}
    \end{minipage}

    \vspace{1em} %
    
    \centering
    \begin{minipage}{0.32\linewidth}
        \centering
        \includegraphics[width=\linewidth]{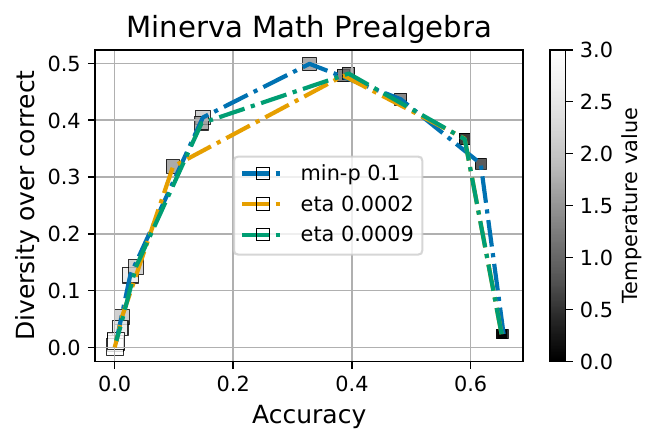}
        \caption*{Minerva $\eta-$sampl.}
    \end{minipage}
    \hfill
    \begin{minipage}{0.32\linewidth}
        \centering
        \includegraphics[width=\linewidth]{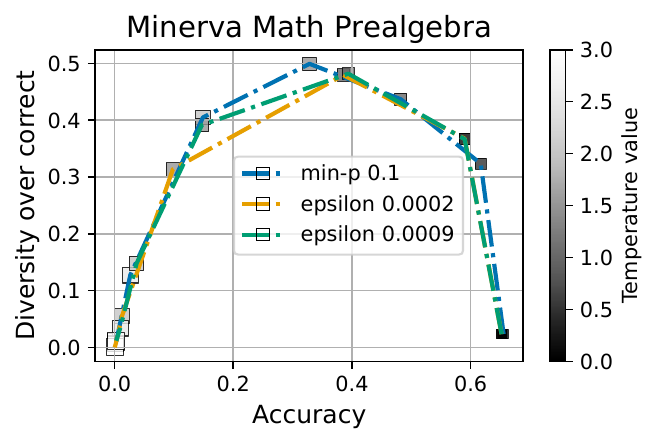}
        \caption*{Minerva $\epsilon-$sampl.}
    \end{minipage}
    \hfill
    \begin{minipage}{0.32\linewidth}
        \centering
        \includegraphics[width=\linewidth]{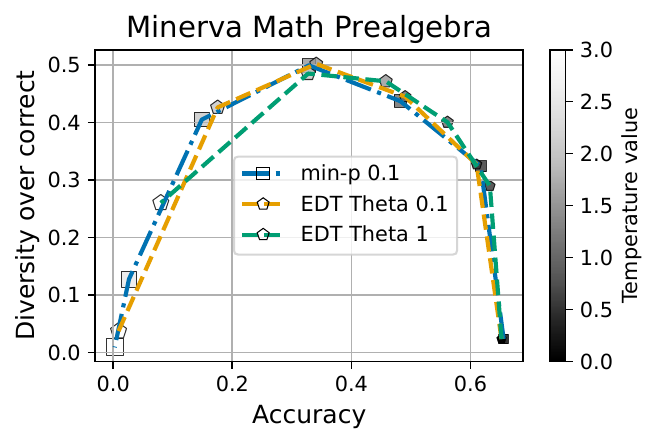}
        \caption*{Minerva EDT}
    \end{minipage}
    \caption{We compare different hyperparameters for the baseline methods on the Minerva task.}
    \label{fig:ablation_minerva}
\end{figure}

%% file: colm2025_conference.bbl
\begin{thebibliography}{62}
\providecommand{\natexlab}[1]{#1}
\providecommand{\url}[1]{\texttt{#1}}
\expandafter\ifx\csname urlstyle\endcsname\relax
  \providecommand{\doi}[1]{doi: #1}\else
  \providecommand{\doi}{doi: \begingroup \urlstyle{rm}\Url}\fi

\bibitem[Ankner et~al.(2025)Ankner, Blakeney, Sreenivasan, Marion, Leavitt, and Paul]{ankner2024perplexed}
Zachary Ankner, Cody Blakeney, Kartik Sreenivasan, Max Marion, Matthew~L Leavitt, and Mansheej Paul.
\newblock Perplexed by perplexity: Perplexity-based data pruning with small reference models.
\newblock In \emph{ICLR}, 2025.
\newblock URL \url{https://openreview.net/forum?id=1GTARJhxtq}.

\bibitem[Ba et~al.(2016)Ba, Kiros, and Hinton]{ba2016layernormalization}
Jimmy~Lei Ba, Jamie~Ryan Kiros, and Geoffrey~E. Hinton.
\newblock Layer normalization.
\newblock \emph{preprint}, 2016.
\newblock URL \url{https://arxiv.org/abs/1607.06450}.

\bibitem[Baan et~al.(2023)Baan, Daheim, Ilia, Ulmer, Li, Fernández, Plank, Sennrich, Zerva, and Aziz]{baan2023uncertaintynaturallanguagegeneration}
Joris Baan, Nico Daheim, Evgenia Ilia, Dennis Ulmer, Haau-Sing Li, Raquel Fernández, Barbara Plank, Rico Sennrich, Chrysoula Zerva, and Wilker Aziz.
\newblock Uncertainty in natural language generation: From theory to applications, 2023.
\newblock URL \url{https://arxiv.org/abs/2307.15703}.

\bibitem[Basu et~al.(2021)Basu, Ramachandran, Keskar, and Varshney]{basu2021mirostat}
Sourya Basu, Govardana~Sachitanandam Ramachandran, Nitish~Shirish Keskar, and Lav~R. Varshney.
\newblock Mirostat: A neural text decoding algorithm that directly controls perplexity.
\newblock In \emph{ICLR}, 2021.
\newblock URL \url{https://openreview.net/forum?id=W1G1JZEIy5_}.

\bibitem[Bubeck \& Cesa-Bianchi(2012)Bubeck and Cesa-Bianchi]{regret_bubeck}
Sébastien Bubeck and Nicolò Cesa-Bianchi.
\newblock Regret analysis of stochastic and nonstochastic multi-armed bandit problems.
\newblock \emph{Foundations and Trends in Machine Learning}, 2012.
\newblock URL \url{http://sbubeck.com/SurveyBCB12.pdf}.

\bibitem[Chang et~al.(2023)Chang, Reitter, Aksitov, and Sung]{chang2023kldivergenceguidedtemperaturesampling}
Chung-Ching Chang, David Reitter, Renat Aksitov, and Yun-Hsuan Sung.
\newblock Kl-divergence guided temperature sampling, 2023.
\newblock URL \url{https://arxiv.org/abs/2306.01286}.

\bibitem[Chen et~al.(2024{\natexlab{a}})Chen, Xiong, Liu, Wu, Xiao, Gao, and He]{chen2024incontext}
Shiqi Chen, Miao Xiong, Junteng Liu, Zhengxuan Wu, Teng Xiao, Siyang Gao, and Junxian He.
\newblock In-context sharpness as alerts: An inner representation perspective for hallucination mitigation.
\newblock In \emph{ICML}, 2024{\natexlab{a}}.
\newblock URL \url{https://openreview.net/forum?id=s3e8poX3kb}.

\bibitem[Chen et~al.(2024{\natexlab{b}})Chen, Jiang, Chang, Hsieh, Yu, and Wang]{chen-etal-2024-minprompt}
Xiusi Chen, Jyun-Yu Jiang, Wei-Cheng Chang, Cho-Jui Hsieh, Hsiang-Fu Yu, and Wei Wang.
\newblock {M}in{P}rompt: Graph-based minimal prompt data augmentation for few-shot question answering.
\newblock In Lun-Wei Ku, Andre Martins, and Vivek Srikumar (eds.), \emph{Proceedings of the 62nd Annual Meeting of the Association for Computational Linguistics (Volume 1: Long Papers)}, pp.\  254--266, Bangkok, Thailand, August 2024{\natexlab{b}}. Association for Computational Linguistics.
\newblock \doi{10.18653/v1/2024.acl-long.16}.
\newblock URL \url{https://aclanthology.org/2024.acl-long.16}.

\bibitem[Chen et~al.(2023)Chen, Fu, Yuan, Wen, Fan, Liu, Zhang, Li, and Xiao]{chen2024hallucinationdetectionrobustlydiscerning}
Yuyan Chen, Qiang Fu, Yichen Yuan, Zhihao Wen, Ge~Fan, Dayiheng Liu, Dongmei Zhang, Zhixu Li, and Yanghua Xiao.
\newblock Hallucination detection: Robustly discerning reliable answers in large language models.
\newblock In \emph{CIKM}, 2023.

\bibitem[Cobbe et~al.(2021)Cobbe, Kosaraju, Bavarian, Chen, Jun, Kaiser, Plappert, Tworek, Hilton, Nakano, Hesse, and Schulman]{cobbe2021gsm8k}
Karl Cobbe, Vineet Kosaraju, Mohammad Bavarian, Mark Chen, Heewoo Jun, Lukasz Kaiser, Matthias Plappert, Jerry Tworek, Jacob Hilton, Reiichiro Nakano, Christopher Hesse, and John Schulman.
\newblock Training verifiers to solve math word problems.
\newblock \emph{arXiv preprint arXiv:2110.14168}, 2021.

\bibitem[DeepSeek-AI et~al.(2025)DeepSeek-AI, Liu, Feng, Xue, Wang, Wu, Lu, Zhao, Deng, Zhang, Ruan, Dai, Guo, Yang, Chen, Ji, Li, Lin, Dai, Luo, Hao, Chen, Li, Zhang, Bao, Xu, Wang, Zhang, Ding, Xin, Gao, Li, Qu, and et~al.]{deepseekai2025deepseekv3technicalreport}
DeepSeek-AI, Aixin Liu, Bei Feng, Bing Xue, Bingxuan Wang, Bochao Wu, Chengda Lu, Chenggang Zhao, Chengqi Deng, Chenyu Zhang, Chong Ruan, Damai Dai, Daya Guo, Dejian Yang, Deli Chen, Dongjie Ji, Erhang Li, Fangyun Lin, Fucong Dai, Fuli Luo, Guangbo Hao, Guanting Chen, Guowei Li, H.~Zhang, Han Bao, Hanwei Xu, Haocheng Wang, Haowei Zhang, Honghui Ding, Huajian Xin, Huazuo Gao, Hui Li, Hui Qu, and J.~L.~Cai et~al.
\newblock Deepseek-v3 technical report, 2025.
\newblock URL \url{https://arxiv.org/abs/2412.19437}.

\bibitem[Deng \& Raffel(2023)Deng and Raffel]{deng-raffel-2023-reward}
Haikang Deng and Colin Raffel.
\newblock Reward-augmented decoding: Efficient controlled text generation with a unidirectional reward model.
\newblock In Houda Bouamor, Juan Pino, and Kalika Bali (eds.), \emph{Proceedings of the 2023 Conference on Empirical Methods in Natural Language Processing}, pp.\  11781--11791, Singapore, December 2023. Association for Computational Linguistics.
\newblock \doi{10.18653/v1/2023.emnlp-main.721}.
\newblock URL \url{https://aclanthology.org/2023.emnlp-main.721}.

\bibitem[Duan et~al.(2024)Duan, Yang, and Tam]{duan2024llmsknowhallucinationempirical}
Hanyu Duan, Yi~Yang, and Kar~Yan Tam.
\newblock Do llms know about hallucination? an empirical investigation of llm's hidden states, 2024.
\newblock URL \url{https://arxiv.org/abs/2402.09733}.

\bibitem[Dubey et~al.(2024)Dubey, Jauhri, Pandey, Kadian, Al-Dahle, Letman, Mathur, Schelten, Yang, Fan, Goyal, Hartshorn, Yang, Mitra, Sravankumar, Korenev, Hinsvark, Rao, Zhang, Rodriguez, Gregerson, Spataru, tiste Roziere, Biron, Tang, Chern, Caucheteux, Nayak, Bi, Marra, McConnell, Keller, Touret, Wu, Wong, Ferrer, Nikolaidis, Allonsius, Song, Pintz, Livshits, Esiobu, Choudhary, Mahajan, Garcia-Olano, Perino, Hupkes, Lakomkin, AlBadawy, Lobanova, and et~al.]{Dubey2024TheL3}
Abhimanyu Dubey, Abhinav Jauhri, Abhinav Pandey, Abhishek Kadian, Ahmad Al-Dahle, Aiesha Letman, Akhil Mathur, Alan Schelten, Amy Yang, Angela Fan, Anirudh Goyal, Anthony~S. Hartshorn, Aobo Yang, Archi Mitra, Archie Sravankumar, Artem Korenev, Arthur Hinsvark, Arun Rao, Aston Zhang, Aur{\'e}lien Rodriguez, Austen Gregerson, Ava Spataru, Bap tiste Roziere, Bethany Biron, Binh Tang, Bobbie Chern, Charlotte Caucheteux, Chaya Nayak, Chloe Bi, Chris Marra, Chris McConnell, Christian Keller, Christophe Touret, Chunyang Wu, Corinne Wong, Cristian~Cant{\'o}n Ferrer, Cyrus Nikolaidis, Damien Allonsius, Daniel Song, Danielle Pintz, Danny Livshits, David Esiobu, Dhruv Choudhary, Dhruv Mahajan, Diego Garcia-Olano, Diego Perino, Dieuwke Hupkes, Egor Lakomkin, Ehab~A. AlBadawy, Elina Lobanova, and Emily~Dinan et~al.
\newblock The llama 3 herd of models.
\newblock \emph{ArXiv}, abs/2407.21783, 2024.
\newblock URL \url{https://api.semanticscholar.org/CorpusID:271571434}.

\bibitem[Fan et~al.(2018)Fan, Lewis, and Dauphin]{fan-etal-2018-hierarchical}
Angela Fan, Mike Lewis, and Yann Dauphin.
\newblock Hierarchical neural story generation.
\newblock In Iryna Gurevych and Yusuke Miyao (eds.), \emph{Proceedings of the 56th Annual Meeting of the Association for Computational Linguistics (Volume 1: Long Papers)}, pp.\  889--898, Melbourne, Australia, July 2018. Association for Computational Linguistics.
\newblock \doi{10.18653/v1/P18-1082}.
\newblock URL \url{https://aclanthology.org/P18-1082}.

\bibitem[Freitag et~al.(2022)Freitag, Grangier, Tan, and Liang]{freitag-etal-2022-high}
Markus Freitag, David Grangier, Qijun Tan, and Bowen Liang.
\newblock High quality rather than high model probability: Minimum {B}ayes risk decoding with neural metrics.
\newblock \emph{Transactions of the Association for Computational Linguistics}, 10:\penalty0 811--825, 2022.
\newblock \doi{10.1162/tacl_a_00491}.
\newblock URL \url{https://aclanthology.org/2022.tacl-1.47}.

\bibitem[Gao et~al.(2024)Gao, Tow, Abbasi, Biderman, Black, DiPofi, Foster, Golding, Hsu, Le~Noac'h, Li, McDonell, Muennighoff, Ociepa, Phang, Reynolds, Schoelkopf, Skowron, Sutawika, Tang, Thite, Wang, Wang, and Zou]{eval-harness}
Leo Gao, Jonathan Tow, Baber Abbasi, Stella Biderman, Sid Black, Anthony DiPofi, Charles Foster, Laurence Golding, Jeffrey Hsu, Alain Le~Noac'h, Haonan Li, Kyle McDonell, Niklas Muennighoff, Chris Ociepa, Jason Phang, Laria Reynolds, Hailey Schoelkopf, Aviya Skowron, Lintang Sutawika, Eric Tang, Anish Thite, Ben Wang, Kevin Wang, and Andy Zou.
\newblock A framework for few-shot language model evaluation, 07 2024.
\newblock URL \url{https://zenodo.org/records/12608602}.

\bibitem[Giulianelli et~al.(2023)Giulianelli, Baan, Aziz, Fern{\'a}ndez, and Plank]{giulianelli-etal-2023-comes}
Mario Giulianelli, Joris Baan, Wilker Aziz, Raquel Fern{\'a}ndez, and Barbara Plank.
\newblock What comes next? evaluating uncertainty in neural text generators against human production variability.
\newblock In Houda Bouamor, Juan Pino, and Kalika Bali (eds.), \emph{Proceedings of the 2023 Conference on Empirical Methods in Natural Language Processing}, pp.\  14349--14371, Singapore, December 2023. Association for Computational Linguistics.
\newblock \doi{10.18653/v1/2023.emnlp-main.887}.
\newblock URL \url{https://aclanthology.org/2023.emnlp-main.887}.

\bibitem[Grattafiori et~al.(2024)Grattafiori, Dubey, Jauhri, and et~al.]{grattafiori2024llama3herdmodels}
Aaron Grattafiori, Abhimanyu Dubey, Abhinav Jauhri, and Abhinav~Pandey et~al.
\newblock The llama 3 herd of models, 2024.
\newblock URL \url{https://arxiv.org/abs/2407.21783}.

\bibitem[Guerreiro et~al.(2023)Guerreiro, Voita, and Martins]{guerreiro-etal-2023-looking}
Nuno~M. Guerreiro, Elena Voita, and Andr{\'e} Martins.
\newblock Looking for a needle in a haystack: A comprehensive study of hallucinations in neural machine translation.
\newblock In Andreas Vlachos and Isabelle Augenstein (eds.), \emph{Proceedings of the 17th Conference of the European Chapter of the Association for Computational Linguistics}, pp.\  1059--1075, Dubrovnik, Croatia, May 2023. Association for Computational Linguistics.
\newblock \doi{10.18653/v1/2023.eacl-main.75}.
\newblock URL \url{https://aclanthology.org/2023.eacl-main.75}.

\bibitem[Hashimoto et~al.(2019)Hashimoto, Zhang, and Liang]{hashimoto-etal-2019-unifying}
Tatsunori~B. Hashimoto, Hugh Zhang, and Percy Liang.
\newblock Unifying human and statistical evaluation for natural language generation.
\newblock In Jill Burstein, Christy Doran, and Thamar Solorio (eds.), \emph{Proceedings of the 2019 Conference of the North {A}merican Chapter of the Association for Computational Linguistics: Human Language Technologies, Volume 1 (Long and Short Papers)}, pp.\  1689--1701, Minneapolis, Minnesota, June 2019. Association for Computational Linguistics.
\newblock \doi{10.18653/v1/N19-1169}.
\newblock URL \url{https://aclanthology.org/N19-1169}.

\bibitem[Hendrycks \& Gimpel(2023)Hendrycks and Gimpel]{hendrycks2023gaussianerrorlinearunits}
Dan Hendrycks and Kevin Gimpel.
\newblock Gaussian error linear units (gelus).
\newblock \emph{preprint}, 2023.
\newblock URL \url{https://arxiv.org/abs/1606.08415}.

\bibitem[Hendrycks et~al.(2021)Hendrycks, Burns, Kadavath, Arora, Basart, Tang, Song, and Steinhardt]{hendrycksmath2021}
Dan Hendrycks, Collin Burns, Saurav Kadavath, Akul Arora, Steven Basart, Eric Tang, Dawn Song, and Jacob Steinhardt.
\newblock Measuring mathematical problem solving with the math dataset.
\newblock \emph{NeurIPS}, 2021.

\bibitem[Hewitt et~al.(2022)Hewitt, Manning, and Liang]{hewitt-etal-2022-truncation}
John Hewitt, Christopher Manning, and Percy Liang.
\newblock Truncation sampling as language model desmoothing.
\newblock In Yoav Goldberg, Zornitsa Kozareva, and Yue Zhang (eds.), \emph{Findings of the Association for Computational Linguistics: EMNLP 2022}, pp.\  3414--3427, Abu Dhabi, United Arab Emirates, December 2022. Association for Computational Linguistics.
\newblock \doi{10.18653/v1/2022.findings-emnlp.249}.
\newblock URL \url{https://aclanthology.org/2022.findings-emnlp.249}.

\bibitem[Holtzman et~al.(2020)Holtzman, Buys, Du, Forbes, and Choi]{Holtzman2020TheCuriousCase}
Ari Holtzman, Jan Buys, Li~Du, Maxwell Forbes, and Yejin Choi.
\newblock The curious case of neural text degeneration.
\newblock In \emph{ICLR}, 2020.
\newblock URL \url{https://openreview.net/forum?id=rygGQyrFvH}.

\bibitem[Jinnai et~al.(2024)Jinnai, Honda, Morimura, and Zhang]{jinnai-etal-2024-generating}
Yuu Jinnai, Ukyo Honda, Tetsuro Morimura, and Peinan Zhang.
\newblock Generating diverse and high-quality texts by minimum {B}ayes risk decoding.
\newblock In Lun-Wei Ku, Andre Martins, and Vivek Srikumar (eds.), \emph{Findings of the Association for Computational Linguistics: ACL 2024}, pp.\  8494--8525, Bangkok, Thailand, August 2024. Association for Computational Linguistics.
\newblock \doi{10.18653/v1/2024.findings-acl.503}.
\newblock URL \url{https://aclanthology.org/2024.findings-acl.503}.

\bibitem[Kingma \& Ba(2015)Kingma and Ba]{kingma:adam}
Diederick~P. Kingma and Jimmy Ba.
\newblock Adam: A method for stochastic optimization.
\newblock In \emph{ICLR}, 2015.
\newblock URL \url{https://arxiv.org/abs/1412.6980}.

\bibitem[Kirk et~al.(2024)Kirk, Mediratta, Nalmpantis, Luketina, Hambro, Grefenstette, and Raileanu]{kirk2024understandingeffectsrlhfllm}
Robert Kirk, Ishita Mediratta, Christoforos Nalmpantis, Jelena Luketina, Eric Hambro, Edward Grefenstette, and Roberta Raileanu.
\newblock Understanding the effects of rlhf on llm generalisation and diversity.
\newblock In \emph{ICLR}, 2024.

\bibitem[Kossen et~al.(2024)Kossen, Han, Razzak, Schut, Malik, and Gal]{Kossen2024SemanticEP}
Jannik Kossen, Jiatong Han, Muhammed Razzak, Lisa Schut, Shreshth~A. Malik, and Yarin Gal.
\newblock Semantic entropy probes: Robust and cheap hallucination detection in llms.
\newblock \emph{ArXiv}, abs/2406.15927, 2024.
\newblock URL \url{https://api.semanticscholar.org/CorpusID:270703114}.

\bibitem[Kumar \& Byrne(2004)Kumar and Byrne]{kumar-byrne-2004-minimum}
Shankar Kumar and William Byrne.
\newblock Minimum {B}ayes-risk decoding for statistical machine translation.
\newblock In \emph{Proceedings of the Human Language Technology Conference of the North {A}merican Chapter of the Association for Computational Linguistics: {HLT}-{NAACL} 2004}, pp.\  169--176, Boston, Massachusetts, USA, May 2 - May 7 2004. Association for Computational Linguistics.
\newblock URL \url{https://aclanthology.org/N04-1022}.

\bibitem[Kwon et~al.(2023)Kwon, Li, Zhuang, Sheng, Zheng, Yu, Gonzalez, Zhang, and Stoica]{kwon2023efficient}
Woosuk Kwon, Zhuohan Li, Siyuan Zhuang, Ying Sheng, Lianmin Zheng, Cody~Hao Yu, Joseph Gonzalez, Hao Zhang, and Ion Stoica.
\newblock Efficient memory management for large language model serving with pagedattention.
\newblock In \emph{Proceedings of the 29th Symposium on Operating Systems Principles}, pp.\  611--626, 2023.

\bibitem[Le~Bronnec et~al.(2024)Le~Bronnec, Verine, Negrevergne, Chevaleyre, and Allauzen]{le-bronnec-etal-2024-exploring}
Florian Le~Bronnec, Alexandre Verine, Benjamin Negrevergne, Yann Chevaleyre, and Alexandre Allauzen.
\newblock Exploring precision and recall to assess the quality and diversity of {LLM}s.
\newblock In Lun-Wei Ku, Andre Martins, and Vivek Srikumar (eds.), \emph{Proceedings of the 62nd Annual Meeting of the Association for Computational Linguistics (Volume 1: Long Papers)}, pp.\  11418--11441, Bangkok, Thailand, August 2024. Association for Computational Linguistics.
\newblock \doi{10.18653/v1/2024.acl-long.616}.
\newblock URL \url{https://aclanthology.org/2024.acl-long.616}.

\bibitem[Li et~al.(2016)Li, Galley, Brockett, Gao, and Dolan]{li-etal-2016-diversity}
Jiwei Li, Michel Galley, Chris Brockett, Jianfeng Gao, and Bill Dolan.
\newblock A diversity-promoting objective function for neural conversation models.
\newblock In Kevin Knight, Ani Nenkova, and Owen Rambow (eds.), \emph{Proceedings of the 2016 Conference of the North {A}merican Chapter of the Association for Computational Linguistics: Human Language Technologies}, pp.\  110--119, San Diego, California, June 2016. Association for Computational Linguistics.
\newblock \doi{10.18653/v1/N16-1014}.
\newblock URL \url{https://aclanthology.org/N16-1014}.

\bibitem[Li et~al.(2023{\natexlab{a}})Li, Patel, Vi{\'e}gas, Pfister, and Wattenberg]{li2023inferencetime}
Kenneth Li, Oam Patel, Fernanda Vi{\'e}gas, Hanspeter Pfister, and Martin Wattenberg.
\newblock Inference-time intervention: Eliciting truthful answers from a language model.
\newblock In \emph{NeurIPS}, 2023{\natexlab{a}}.
\newblock URL \url{https://openreview.net/forum?id=aLLuYpn83y}.

\bibitem[Li et~al.(2023{\natexlab{b}})Li, Holtzman, Fried, Liang, Eisner, Hashimoto, Zettlemoyer, and Lewis]{li-etal-2023-contrastive}
Xiang~Lisa Li, Ari Holtzman, Daniel Fried, Percy Liang, Jason Eisner, Tatsunori Hashimoto, Luke Zettlemoyer, and Mike Lewis.
\newblock Contrastive decoding: Open-ended text generation as optimization.
\newblock In Anna Rogers, Jordan Boyd-Graber, and Naoaki Okazaki (eds.), \emph{Proceedings of the 61st Annual Meeting of the Association for Computational Linguistics (Volume 1: Long Papers)}, pp.\  12286--12312, Toronto, Canada, July 2023{\natexlab{b}}. Association for Computational Linguistics.
\newblock \doi{10.18653/v1/2023.acl-long.687}.
\newblock URL \url{https://aclanthology.org/2023.acl-long.687}.

\bibitem[Liu et~al.(2024)Liu, Frawley, Wyer, Shum, Uckelman, Black, and Willcocks]{liu-etal-2024-self-regulated}
Mingyue Liu, Jonathan Frawley, Sarah Wyer, Hubert P.~H. Shum, Sara Uckelman, Sue Black, and Chris Willcocks.
\newblock Self-regulated sample diversity in large language models.
\newblock In Kevin Duh, Helena Gomez, and Steven Bethard (eds.), \emph{Findings of the Association for Computational Linguistics: NAACL 2024}, pp.\  1891--1899, Mexico City, Mexico, June 2024. Association for Computational Linguistics.
\newblock \doi{10.18653/v1/2024.findings-naacl.122}.
\newblock URL \url{https://aclanthology.org/2024.findings-naacl.122}.

\bibitem[Mahaut et~al.(2024)Mahaut, Aina, Czarnowska, Hardalov, M{\"u}ller, and Marquez]{mahaut-etal-2024-factual}
Mat{\'e}o Mahaut, Laura Aina, Paula Czarnowska, Momchil Hardalov, Thomas M{\"u}ller, and Lluis Marquez.
\newblock Factual confidence of {LLM}s: on reliability and robustness of current estimators.
\newblock In Lun-Wei Ku, Andre Martins, and Vivek Srikumar (eds.), \emph{Proceedings of the 62nd Annual Meeting of the Association for Computational Linguistics (Volume 1: Long Papers)}, pp.\  4554--4570, Bangkok, Thailand, August 2024. Association for Computational Linguistics.
\newblock \doi{10.18653/v1/2024.acl-long.250}.
\newblock URL \url{https://aclanthology.org/2024.acl-long.250}.

\bibitem[Marion et~al.(2023)Marion, {\"U}st{\"u}n, Pozzobon, Wang, Fadaee, and Hooker]{marion2023less}
Max Marion, Ahmet {\"U}st{\"u}n, Luiza Pozzobon, Alex Wang, Marzieh Fadaee, and Sara Hooker.
\newblock When less is more: Investigating data pruning for pretraining llms at scale.
\newblock In \emph{NeurIPS Workshop on Attributing Model Behavior at Scale (ATTRIB)}, 2023.
\newblock URL \url{https://openreview.net/forum?id=XUIYn3jo5T}.

\bibitem[Meister et~al.(2023)Meister, Pimentel, Malagutti, Wilcox, and Cotterell]{meister-etal-2023-efficacy}
Clara Meister, Tiago Pimentel, Luca Malagutti, Ethan Wilcox, and Ryan Cotterell.
\newblock On the efficacy of sampling adapters.
\newblock In Anna Rogers, Jordan Boyd-Graber, and Naoaki Okazaki (eds.), \emph{Proceedings of the 61st Annual Meeting of the Association for Computational Linguistics (Volume 1: Long Papers)}, pp.\  1437--1455, Toronto, Canada, July 2023. Association for Computational Linguistics.
\newblock \doi{10.18653/v1/2023.acl-long.80}.
\newblock URL \url{https://aclanthology.org/2023.acl-long.80}.

\bibitem[Meng et~al.(2025)Meng, Wu, and Monz]{meng-etal-2025-learn}
Yan Meng, Di~Wu, and Christof Monz.
\newblock How to learn in a noisy world? self-correcting the real-world data noise in machine translation.
\newblock In Luis Chiruzzo, Alan Ritter, and Lu~Wang (eds.), \emph{Findings of the Association for Computational Linguistics: NAACL 2025}, pp.\  7451--7467, Albuquerque, New Mexico, April 2025. Association for Computational Linguistics.
\newblock ISBN 979-8-89176-195-7.
\newblock \doi{10.18653/v1/2025.findings-naacl.416}.
\newblock URL \url{https://aclanthology.org/2025.findings-naacl.416/}.

\bibitem[Minh et~al.(2025)Minh, Baker, Neo, Roush, Kirsch, and Shwartz-Ziv]{nguyen2024turningheatminpsampling}
Nguyen~Nhat Minh, Andrew Baker, Clement Neo, Allen~G Roush, Andreas Kirsch, and Ravid Shwartz-Ziv.
\newblock Turning up the heat: Min-p sampling for creative and coherent {LLM} outputs.
\newblock In \emph{ICLR}, 2025.
\newblock URL \url{https://openreview.net/forum?id=FBkpCyujtS}.

\bibitem[Mirzadeh et~al.(2025)Mirzadeh, Alizadeh, Shahrokhi, Tuzel, Bengio, and Farajtabar]{mirzadeh2025gsmsymbolic}
Seyed~Iman Mirzadeh, Keivan Alizadeh, Hooman Shahrokhi, Oncel Tuzel, Samy Bengio, and Mehrdad Farajtabar.
\newblock {GSM}-symbolic: Understanding the limitations of mathematical reasoning in large language models.
\newblock In \emph{ICLR}, 2025.
\newblock URL \url{https://openreview.net/forum?id=AjXkRZIvjB}.

\bibitem[Mudgal et~al.(2024)Mudgal, Lee, Ganapathy, Li, Wang, Huang, Chen, Cheng, Collins, Strohman, Chen, Beutel, and Beirami]{mudgal2024controlled}
Sidharth Mudgal, Jong Lee, Harish Ganapathy, YaGuang Li, Tao Wang, Yanping Huang, Zhifeng Chen, Heng-Tze Cheng, Michael Collins, Trevor Strohman, Jilin Chen, Alex Beutel, and Ahmad Beirami.
\newblock Controlled decoding from language models.
\newblock In \emph{ICML}, 2024.
\newblock URL \url{https://openreview.net/forum?id=bVIcZb7Qa0}.

\bibitem[OpenAI et~al.(2024)OpenAI, Achiam, Adler, Agarwal, Ahmad, Akkaya, Aleman, Almeida, Altenschmidt, Altman, Anadkat, Avila, Babuschkin, Balaji, Balcom, Baltescu, Bao, Bavarian, Belgum, Bello, Berdine, Bernadett-Shapiro, Berner, Bogdonoff, Boiko, Boyd, Brakman, Brockman, Brooks, Brundage, Button, Cai, Campbell, Cann, Carey, Carlson, Carmichael, Chan, Chang, Chantzis, Chen, Chen, and et~al.]{openai2024gpt4technicalreport}
OpenAI, Josh Achiam, Steven Adler, Sandhini Agarwal, Lama Ahmad, Ilge Akkaya, Florencia~Leoni Aleman, Diogo Almeida, Janko Altenschmidt, Sam Altman, Shyamal Anadkat, Red Avila, Igor Babuschkin, Suchir Balaji, Valerie Balcom, Paul Baltescu, Haiming Bao, Mohammad Bavarian, Jeff Belgum, Irwan Bello, Jake Berdine, Gabriel Bernadett-Shapiro, Christopher Berner, Lenny Bogdonoff, Oleg Boiko, Madelaine Boyd, Anna-Luisa Brakman, Greg Brockman, Tim Brooks, Miles Brundage, Kevin Button, Trevor Cai, Rosie Campbell, Andrew Cann, Brittany Carey, Chelsea Carlson, Rory Carmichael, Brooke Chan, Che Chang, Fotis Chantzis, Derek Chen, Sully Chen, and et~al.
\newblock Gpt-4 technical report, 2024.
\newblock URL \url{https://arxiv.org/abs/2303.08774}.

\bibitem[Pedregosa et~al.(2011)Pedregosa, Varoquaux, Gramfort, Michel, Thirion, Grisel, Blondel, Prettenhofer, Weiss, Dubourg, et~al.]{pedregosa2011scikit}
Fabian Pedregosa, Ga{\"e}l Varoquaux, Alexandre Gramfort, Vincent Michel, Bertrand Thirion, Olivier Grisel, Mathieu Blondel, Peter Prettenhofer, Ron Weiss, Vincent Dubourg, et~al.
\newblock Scikit-learn: Machine learning in python.
\newblock \emph{JMLR}, 12\penalty0 (Oct):\penalty0 2825--2830, 2011.

\bibitem[Renze(2024)]{renze-2024-effect}
Matthew Renze.
\newblock The effect of sampling temperature on problem solving in large language models.
\newblock In Yaser Al-Onaizan, Mohit Bansal, and Yun-Nung Chen (eds.), \emph{Findings of the Association for Computational Linguistics: EMNLP 2024}, pp.\  7346--7356, Miami, Florida, USA, November 2024. Association for Computational Linguistics.
\newblock URL \url{https://aclanthology.org/2024.findings-emnlp.432}.

\bibitem[Shannon(1948)]{shannon1948mathematical}
Claude~E Shannon.
\newblock A mathematical theory of communication.
\newblock \emph{The Bell System Technical Journal}, 27\penalty0 (3):\penalty0 379--423, 1948.

\bibitem[Shi et~al.(2024)Shi, Yang, Cai, Zhang, Wang, Yang, and Lam]{shi-etal-2024-thorough}
Chufan Shi, Haoran Yang, Deng Cai, Zhisong Zhang, Yifan Wang, Yujiu Yang, and Wai Lam.
\newblock A thorough examination of decoding methods in the era of {LLM}s.
\newblock In Yaser Al-Onaizan, Mohit Bansal, and Yun-Nung Chen (eds.), \emph{Proceedings of the 2024 Conference on Empirical Methods in Natural Language Processing}, pp.\  8601--8629, Miami, Florida, USA, November 2024. Association for Computational Linguistics.
\newblock URL \url{https://aclanthology.org/2024.emnlp-main.489}.

\bibitem[Stiennon et~al.(2020)Stiennon, Ouyang, Wu, Ziegler, Lowe, Voss, Radford, Amodei, and Christiano]{bestofn_summarization}
Nisan Stiennon, Long Ouyang, Jeff Wu, Daniel~M. Ziegler, Ryan Lowe, Chelsea Voss, Alec Radford, Dario Amodei, and Paul Christiano.
\newblock Learning to summarize from human feedback.
\newblock In \emph{NeurIPS}, 2020.
\newblock ISBN 9781713829546.

\bibitem[Suzgun et~al.(2023)Suzgun, Melas-Kyriazi, and Jurafsky]{suzgun-etal-2023-follow}
Mirac Suzgun, Luke Melas-Kyriazi, and Dan Jurafsky.
\newblock Follow the wisdom of the crowd: Effective text generation via minimum {B}ayes risk decoding.
\newblock In Anna Rogers, Jordan Boyd-Graber, and Naoaki Okazaki (eds.), \emph{Findings of the Association for Computational Linguistics: ACL 2023}, pp.\  4265--4293, Toronto, Canada, July 2023. Association for Computational Linguistics.
\newblock \doi{10.18653/v1/2023.findings-acl.262}.
\newblock URL \url{https://aclanthology.org/2023.findings-acl.262}.

\bibitem[Troshin et~al.(2024)Troshin, Niculae, and Fokkens]{troshin2024efficientcontrolledlanguagegeneration}
Sergey Troshin, Vlad Niculae, and Antske Fokkens.
\newblock Efficient controlled language generation with low-rank autoregressive reward models, 2024.
\newblock URL \url{https://arxiv.org/abs/2407.04615}.

\bibitem[Wang \& Zhou(2024)Wang and Zhou]{wang2024chainofthought}
Xuezhi Wang and Denny Zhou.
\newblock Chain-of-thought reasoning without prompting.
\newblock In \emph{NeurIPS}, 2024.
\newblock URL \url{https://openreview.net/forum?id=4Zt7S0B0Jp}.

\bibitem[Wang et~al.(2024)Wang, Ma, Zhang, Ni, Chandra, Guo, Ren, Arulraj, He, Jiang, Li, Ku, Wang, Zhuang, Fan, Yue, and Chen]{wang2024mmlupro}
Yubo Wang, Xueguang Ma, Ge~Zhang, Yuansheng Ni, Abhranil Chandra, Shiguang Guo, Weiming Ren, Aaran Arulraj, Xuan He, Ziyan Jiang, Tianle Li, Max Ku, Kai Wang, Alex Zhuang, Rongqi Fan, Xiang Yue, and Wenhu Chen.
\newblock {MMLU}-pro: A more robust and challenging multi-task language understanding benchmark.
\newblock In \emph{The Thirty-eight Conference on Neural Information Processing Systems Datasets and Benchmarks Track}, 2024.
\newblock URL \url{https://openreview.net/forum?id=y10DM6R2r3}.

\bibitem[Wolf et~al.(2020)Wolf, Debut, Sanh, Chaumond, Delangue, Moi, Cistac, Rault, Louf, Funtowicz, Davison, Shleifer, von Platen, Ma, Jernite, Plu, Xu, Scao, Gugger, Drame, Lhoest, and Rush]{wolf2020huggingfacestransformersstateoftheartnatural}
Thomas Wolf, Lysandre Debut, Victor Sanh, Julien Chaumond, Clement Delangue, Anthony Moi, Pierric Cistac, Tim Rault, Rémi Louf, Morgan Funtowicz, Joe Davison, Sam Shleifer, Patrick von Platen, Clara Ma, Yacine Jernite, Julien Plu, Canwen Xu, Teven~Le Scao, Sylvain Gugger, Mariama Drame, Quentin Lhoest, and Alexander~M. Rush.
\newblock Huggingface's transformers: State-of-the-art natural language processing, 2020.
\newblock URL \url{https://arxiv.org/abs/1910.03771}.

\bibitem[Wu et~al.(2025)Wu, Fernandes, Bertsch, Kim, Pakazad, and Neubig]{instruction-following-mbr}
Ian Wu, Patrick Fernandes, Amanda Bertsch, Seungone Kim, Sina~Khoshfetrat Pakazad, and Graham Neubig.
\newblock Better instruction-following through minimum bayes risk.
\newblock In \emph{ICLR}, 2025.
\newblock URL \url{https://openreview.net/forum?id=7xCSK9BLPy}.

\bibitem[Yang et~al.(2024)Yang, Yang, Zhang, Hui, Zheng, Yu, Li, Liu, Huang, Dong, Wei, Lin, Yang, Tu, Zhang, Yang, Yang, Zhou, Lin, Dang, Lu, Bao, Yang, Yu, Li, Xue, Zhang, Zhu, Men, Lin, Li, Xia, Ren, Ren, Fan, Su, Zhang, Wan, Liu, Cui, Zhang, Qiu, Quan, and Wang]{Yang2024Qwen25TR}
Qwen~An Yang, Baosong Yang, Beichen Zhang, Binyuan Hui, Bo~Zheng, Bowen Yu, Chengyuan Li, Dayiheng Liu, Fei Huang, Guanting Dong, Haoran Wei, Huan Lin, Jian Yang, Jianhong Tu, Jianwei Zhang, Jianxin Yang, Jiaxin Yang, Jingren Zhou, Junyang Lin, Kai Dang, Keming Lu, Keqin Bao, Kexin Yang, Le~Yu, Mei Li, Mingfeng Xue, Pei Zhang, Qin Zhu, Rui Men, Runji Lin, Tianhao Li, Tingyu Xia, Xingzhang Ren, Xuancheng Ren, Yang Fan, Yang Su, Yi-Chao Zhang, Yunyang Wan, Yuqi Liu, Zeyu Cui, Zhenru Zhang, Zihan Qiu, Shanghaoran Quan, and Zekun Wang.
\newblock Qwen2.5 technical report.
\newblock \emph{ArXiv}, abs/2412.15115, 2024.
\newblock URL \url{https://api.semanticscholar.org/CorpusID:274859421}.

\bibitem[Yao et~al.(2023)Yao, Yu, Zhao, Shafran, Griffiths, Cao, and Narasimhan]{shunyu_tree_of_thoughts}
Shunyu Yao, Dian Yu, Jeffrey Zhao, Izhak Shafran, Thomas~L. Griffiths, Yuan Cao, and Karthik Narasimhan.
\newblock Tree of thoughts: deliberate problem solving with large language models.
\newblock In \emph{NeurIPS}, 2023.

\bibitem[Yuan et~al.(2023)Yuan, Yuan, Li, Dong, Lu, Tan, Zhou, and Zhou]{yuan2023scaling}
Zheng Yuan, Hongyi Yuan, Chengpeng Li, Guanting Dong, Keming Lu, Chuanqi Tan, Chang Zhou, and Jingren Zhou.
\newblock Scaling relationship on learning mathematical reasoning with large language models, 2023.
\newblock URL \url{https://arxiv.org/abs/2308.01825}.

\bibitem[Zhang et~al.(2024{\natexlab{a}})Zhang, Jiang, Shu, Zheng, Wei, et~al.]{zhang2024noise}
Feipeng Zhang, Wenyu Jiang, Jun Shu, Feng Zheng, Hongxin Wei, et~al.
\newblock On the noise robustness of in-context learning for text generation.
\newblock \emph{NeurIPS}, 2024{\natexlab{a}}.

\bibitem[Zhang et~al.(2021)Zhang, Duckworth, Ippolito, and Neelakantan]{zhang-etal-2021-trading}
Hugh Zhang, Daniel Duckworth, Daphne Ippolito, and Arvind Neelakantan.
\newblock Trading off diversity and quality in natural language generation.
\newblock In Anya Belz, Shubham Agarwal, Yvette Graham, Ehud Reiter, and Anastasia Shimorina (eds.), \emph{Proceedings of the Workshop on Human Evaluation of NLP Systems (HumEval)}, pp.\  25--33, Online, April 2021. Association for Computational Linguistics.
\newblock URL \url{https://aclanthology.org/2021.humeval-1.3}.

\bibitem[Zhang et~al.(2024{\natexlab{b}})Zhang, Bao, and Huang]{zhang2024edtimprovinglargelanguage}
Shimao Zhang, Yu~Bao, and Shujian Huang.
\newblock {EDT}: Improving large language models' generation by entropy-based dynamic temperature sampling, 2024{\natexlab{b}}.
\newblock URL \url{https://arxiv.org/abs/2403.14541}.

\bibitem[Zhu et~al.(2024)Zhu, Li, Li, Zhao, Jin, and Mei]{zhu2023hotcoldadaptivetemperature}
Yuqi Zhu, Jia Li, Ge~Li, YunFei Zhao, Zhi Jin, and Hong Mei.
\newblock Hot or cold? adaptive temperature sampling for code generation with large language models.
\newblock In \emph{AAAI}, 2024.

\end{thebibliography}
